\documentclass{article} % For LaTeX2e
\usepackage{iclr2023_conference,times}

\usepackage{amsmath,amsfonts,bm}

\def\eqref#1{equation~\ref{#1}}
\def\1{\bm{1}}

\DeclareMathAlphabet{\mathsfit}{\encodingdefault}{\sfdefault}{m}{sl}
\SetMathAlphabet{\mathsfit}{bold}{\encodingdefault}{\sfdefault}{bx}{n}

\usepackage{hyperref}
\usepackage{url}

\usepackage{booktabs}       % professional-quality tables
\usepackage{amsfonts}       % blackboard math symbols
\usepackage{nicefrac}       % compact symbols for 1/2, etc.
\usepackage{microtype}      % microtypography
\usepackage{xcolor}         % colors

\usepackage{graphicx}
\usepackage{algorithm}
\usepackage{algorithmic}
\usepackage{amsmath}
\usepackage{subcaption}
\usepackage{multirow,tabularx}
\usepackage{makecell}
\usepackage{array}
\usepackage{eqparbox}
\usepackage{wrapfig}
\usepackage{caption}
\usepackage{pifont}

\newcommand{\red}[1]{\textcolor{black}{#1}}

\title{Training-Free Structured Diffusion Guidance for Compositional Text-to-Image Synthesis}

\author{Weixi Feng$^1$, Xuehai He$^2$, Tsu-jui Fu$^1$, Varun Jampani$^3$, Arjun Akula$^3$, \\ \textbf{Pradyumna Narayana$^3$, Sugato Basu$^3$, Xin Eric Wang$^2$, William Yang Wang$^1$} \\
$^1$University of California, Santa Barbara, $^2$University of California, Santa Cruz, $^3$Google\\
}

\iclrfinalcopy % Uncomment for camera-ready version, but NOT for submission.
\begin{document}

\maketitle

\begin{abstract}
Large-scale diffusion models have achieved state-of-the-art results on text-to-image synthesis (T2I) tasks. Despite their ability to generate high-quality yet creative images, we observe that attribution-binding and compositional capabilities are still considered major challenging issues, especially when involving multiple objects. \red{Attribute-binding requires the model to associate objects with the correct attribute descriptions, and compositional skills require the model to combine and generate multiple concepts into a single image. 
In this work, we improve these two aspects of T2I models to achieve more accurate image compositions.}
To do this, we incorporate linguistic structures with the diffusion guidance process based on the controllable properties of manipulating cross-attention layers in diffusion-based T2I models. 
We observe that keys and values in cross-attention layers have strong semantic meanings associated with object layouts and content. Therefore, by manipulating the cross-attention representations based on linguistic insights, we can better preserve the compositional semantics in the generated image.
Built upon Stable Diffusion, a SOTA T2I model, our structured cross-attention design is efficient that requires no additional training samples.
We achieve better compositional skills in qualitative and quantitative results, leading to a significant 5-8\% advantage in head-to-head user comparison studies.
Lastly, we conduct an in-depth analysis to reveal potential causes of incorrect image compositions and justify the properties of cross-attention layers in the generation process.

\end{abstract}

\section{Introduction} \label{sec:label}

Text-to-Image Synthesis (T2I) is to generate natural and faithful images given a text prompt as input. Recently, there has been a significant advancement in the quality of generated images by extremely large-scale vision-language models, such as DALL-E 2 \citep{dalle2}, Imagen \citep{imagen}, and Parti \citep{parti}. In particular, Stable Diffusion \citep{rombach2022high} is the state-of-the-art open-source implementation showing superior evaluation metric gains after training over billions of text-image pairs.

In addition to generating high-fidelity images, the ability to compose multiple objects into a coherent scene is also essential. \red{Given a text prompt from the user end, T2I models need to generate an image that contains all necessary visual concepts as mentioned in the text.}
Achieving such ability requires the model to understand both the full prompt and individual linguistic concepts from the prompt. As a result, the model should be able to combine multiple concepts and generate novel objects that have never been included in the training data. In this work, we mainly focus on improving the compositionality of the generation process, as it is essential to achieve controllable and generalized text-to-image synthesis with multiple objects in a complex scene. 

\begin{figure*}[t]
    \centering
    \includegraphics[width=0.9\textwidth]{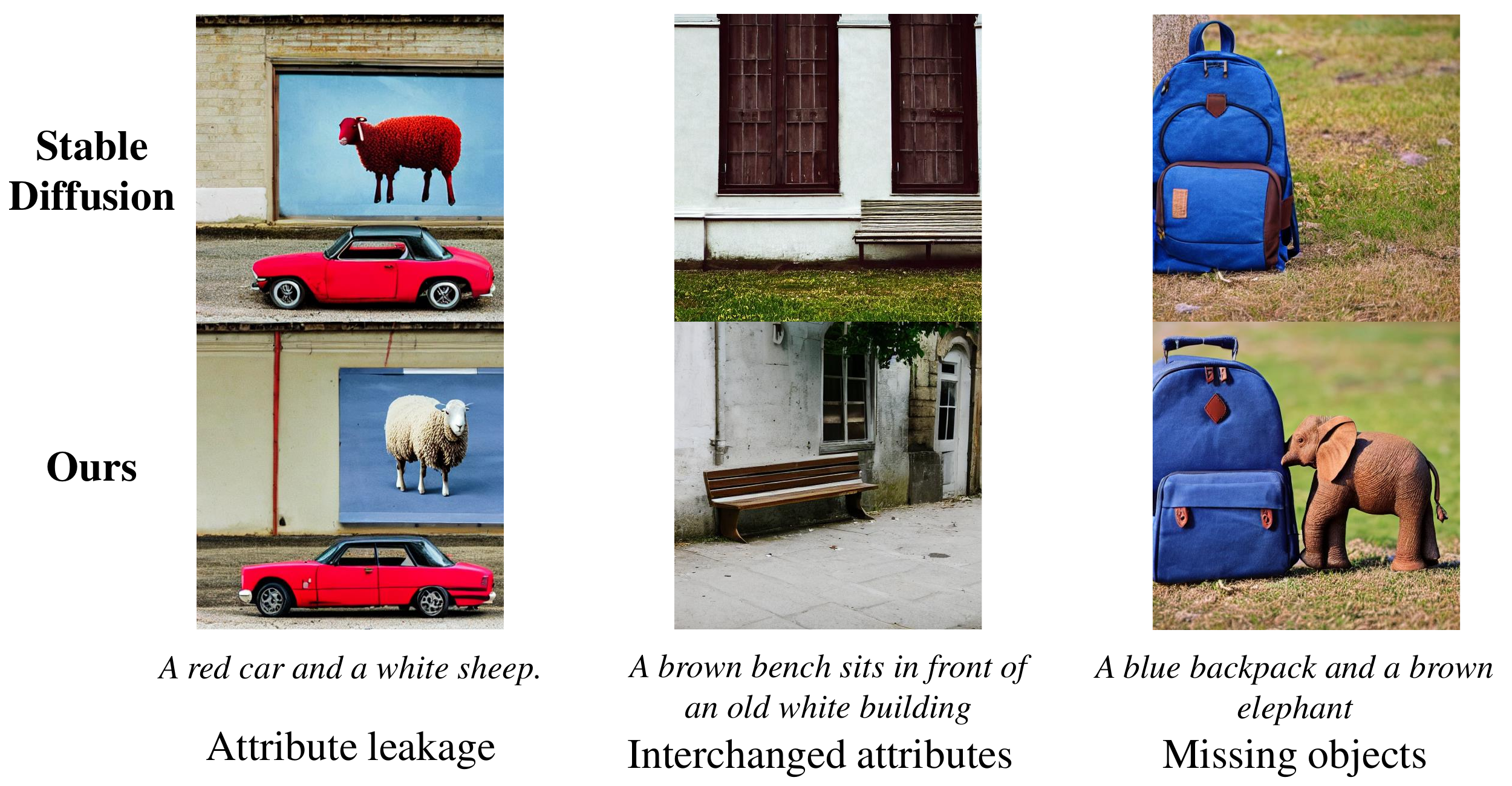}
    \caption{\textbf{Three challenging phenomena in the compositional generation.} Attribute leakage: The attribute of one object is (partially) observable in another object. Interchanged attributes: the attributes of two or more objects are interchanged. Missing objects: one or more objects are missing. With slight abuse of attribute binding definitions, we aim to address all three problems in this work. 
    }
    \label{fig:teaser}
\end{figure*}

\textit{Attribute binding} is a critical compositionality challenge \citep{dalle2, imagen} to existing large-scale diffusion-based models. Despite the improvements in generating multiple objects in the same scene, existing models still fail when given a prompt such as ``a  brown bench in front of a white building'' (see Fig. \ref{fig:teaser}). The output images contains ``a white bench'' and ``a brown building'' instead, potentially due to strong training set bias or imprecise language understanding. From a practical perspective, explaining and solving such a two-object binding challenge is a primary step to understanding more complex prompts with multiple objects. Therefore, how to bind the attributes to the correct objects is a fundamental problem for a more complicated and reliable compositional generation. While previous work has addressed compositional T2I \citep{park2021benchmark}, our work tackles open-domain foreground objects with counterfactual attributes, such as color and materials.

Even though state-of-the-art (SOTA) T2I models are trained on large-scale text-image datasets, they can still suffer from inaccurate results for simple prompts similar to the example above. Hence, we are motivated to seek an alternative, data-efficient method to improve the compositionality. We observe that the attribute-object relation pairs can be obtained as text spans for free from the parsing tree of the sentence. Therefore, we propose to combine the structured representations of prompts, such as a constituency tree or a scene graph, with the diffusion guidance process. Text spans only depict limited regions of the whole image. Conventionally, we need spatial information such as coordinates \citep{yang2022modeling} as input to map their semantics into corresponding images. However, coordinate inputs cannot be interpreted by T2I models. Instead, we make use of the observations that attention maps provide free \textit{token-region associations} in trained T2I models \citep{hertz2022prompt}. By modifying the key-value pairs in cross-attention layers, we manage to map the encoding of each text span into attended regions in 2D image space.

In this work, we discover similar observations in Stable Diffusion \citep{rombach2022high} and utilize the property to build structured cross-attention guidance. Specifically, we use language parsers to obtain hierarchical structures from the prompts. We extract text spans across all levels, including visual concepts or entities, and encode them separately to disentangle the attribute-object pairs from each other. 
Compared to using a single sequence of text embedding for guidance, we improve the compositionality by multiple sequences where each emphasizes an entity or a union of entities from multiple hierarchies in the structured language representations. We refer to our method as Structured Diffusion Guidance (StructureDiffusion). Our contributions can be summarized as three-fold:

\begin{itemize}
    \item We propose an intuitive and effective method to improve compositional text-to-image synthesis by utilizing structured representations of language inputs. Our method is efficient and training-free that requires no additional training samples.
    \item Experimental results show that our method achieves more accurate attribute binding and compositionality in the generated images. We also propose a benchmark named \textbf{A}ttribute \textbf{B}inding \textbf{C}ontrast set (ABC-6K) to measure the compositional skills of T2I models. 
    \item We conduct extensive experiments and analysis to identify the causes of incorrect attribute binding, which points out future directions in improving the faithfulness and compositionality of text-to-image synthesis. 
\end{itemize}

\section{Diffusion Models \& Structured Guidance} \label{sec:method}

\begin{figure*}[t]
    \centering
    \includegraphics[width=\textwidth]{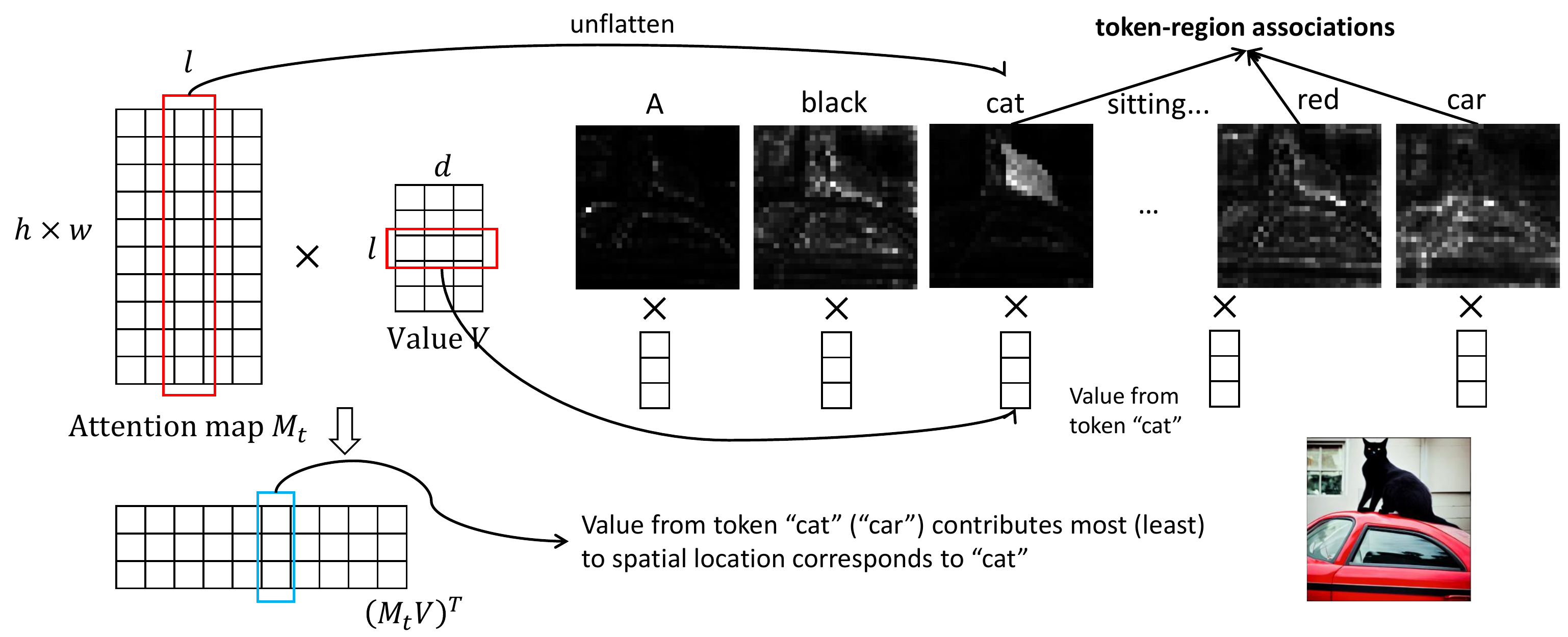}
    \caption{\red{An illustration of cross-attention operations and the token-region associations from attention maps. We omit some tokens for simplicity.}}
    \label{fig:attention}
\end{figure*}

In this section, we propose a simple yet effective approach incorporating structured language representations into the cross-attention layers. We briefly introduce the Stable Diffusion model and its critical components in Sec. \ref{subsec:m1}. Then, we present our method in detail in Sec. \ref{subsec:m2}.

\subsection{Background} \label{subsec:m1}
\paragraph{Stable Diffusion} We implement our approach and experiments on the state-of-the-art T2I model, Stable Diffusion \citep{rombach2022high}. It is a two-stage method that consists of an autoencoder and a diffusion model. The pre-trained autoencoder encodes images as lower-resolution latent maps for diffusion training. During inference, it decodes generated outputs from the diffusion model into images. The diffusion model generates lower-resolution latent maps based on a random Gaussian noise input $z^{T}$. Given $z^{T}$, it outputs a noise estimation $\epsilon$ at each step $t$ and subtracts it from $z^{t}$. The final noise-free latent map prediction $z^0$ is fed into the autoencoder to generate images.  
Stable Diffusion adopts a modified UNet \citep{unet} for noise estimation and a frozen \textit{CLIP text encoder} \citep{clip} to encode text inputs as embedding sequences. The interactions between the image space and the textual embeddings are achieved through multiple \textit{cross-attention layers} in both \red{downsampling} and upsampling blocks.  

\paragraph{CLIP Text Encoder} Given an input prompt \red{$\mathcal{P}$}, the CLIP encoder encodes it as a sequence of embeddings \red{$\mathcal{W}_{\text{p}}=\text{CLIP}_{\text{text}}(\mathcal{P})$} where \red{$c_{\text{p}}$} is the embedding dimension and $l$ is the sequence length. Our key observation is that the contextualization of CLIP embeddings is a potential cause of incorrect attribute binding. Due to the causal attention masks, tokens in the later part of a sequence are blended with the token semantics before them. For example, When the user indicates some rare color for the second object (e.g. ``a yellow apple and red bananas''), Stable Diffusion tends to generate ``banana'' in ``yellow'', as the embeddings of ``yellow'' is attended by token ``banana''. 

\paragraph{Cross Attention Layers} The cross-attention layers take the embedding sequences from the CLIP text encoder and fuse them with latent feature maps to achieve classifier-free guidance. Denote a 2D feature map $\mathcal{X}^{t}$, it is projected into queries by a linear layer $f_{Q}(\cdot)$ and reshaped as $Q^{t} \in R^{(n, h \times w, d)}$ where $n$ denotes the number of attention heads, $d$ is the feature dimension. Similarly $\mathcal{W}_{\text{p}}$ is projected as keys and values \red{$K_{\text{p}}, V_{\text{p}} \in R^{(n, l, d)}$} by linear layers $f_{K}(\cdot), f_{V}(\cdot)$. The attention maps refer to the product between queries and keys, denoted as a function $f_{M}(\cdot)$
\begin{equation} \label{eq:attn_map}
    M^{t} = f_{M}(Q^t, K_{p}) = \text{Softmax}(\frac{Q^{t}K_{\text{p}}^T}{\sqrt{d}}), \; M^{t} \in R^{(n, h\times w, l)}.
\end{equation}

\begin{figure*}[t]
    \centering
    \includegraphics[width=0.95\textwidth]{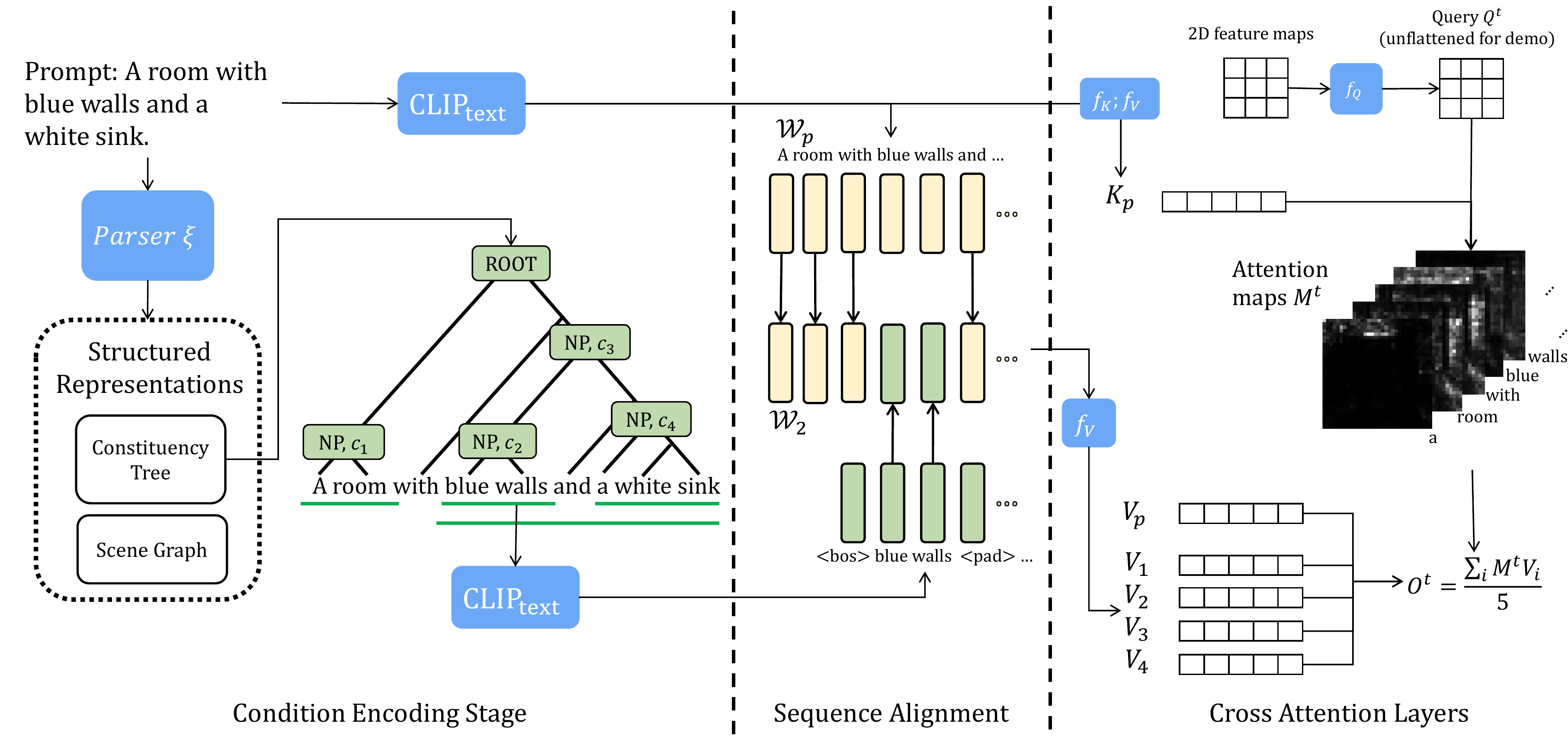}
    \caption{An illustration of our cross-attention design with structured representations. We unflatten the query and attention maps and omit the feature dimension $d$ of all query, key, and value tensors for demonstration purposes. Note that noun phrases at multiple hierarchies are extracted and encoded through the frozen CLIP text encoder and projected to value vectors. }
    \label{fig:method}
\end{figure*}

\paragraph{Cross Attention Controls} \red{\citet{hertz2022prompt} observes that the spatial layouts depend on the cross attention maps in Imagen \cite{imagen}. These maps control the layout and structure of generated images, while the values contain rich semantics mapped into attended regions. Therefore, we assume that the image layout and content can be disentangled by controlling attention maps and values separately. }

\subsection{Structured Diffusion Guidance} \label{subsec:m2}
Given the challenging prompts in Fig. \ref{fig:teaser}, the attribute-object pairs are available \textbf{for free}\footnote{For free means that the extra computational cost introduced here is trivial in the whole diffusion process.} in many structured representations, such as a constituency tree or a scene graph. We seek an implicit way of combining language structures with the cross-attention layers. \red{As is shown in Fig. \ref{fig:method}, we can extract multiple noun phrases (NPs) and map their semantics into corresponding regions. Since $M_t$ provides natural token-region associations (see Fig. \ref{fig:attention}), we can apply it to multiple values from different NPs to achieve region-wise semantic guidance.}

Specifically, given a parser $\xi(\cdot)$, we first extract a collection of concepts from all hierarchical levels as $\mathcal{C}=\{c_1, c_2,\ldots,c_k\}$. For constituency parsing, we extract all NPs from the tree structure (see Fig.\ref{fig:method} left). For the scene graphs, we extract objects and their relations with another object as text segments. We encode each NP separately:
\begin{equation}
\begin{split}
    \mathbb{W} = [\red{\mathcal{W}_{\text{p}},} \mathcal{W}_1,\mathcal{W}_2,\ldots,\mathcal{W}_k],\; \mathcal{W}_i=\text{CLIP}_{\text{text}}(c_i), \; i=1,\ldots k.
\end{split}
\end{equation}
\red{The embedding sequence $\mathcal{W}_i$ is realigned with $\mathcal{W}_p$ as shown in the middle of Fig. \ref{fig:method}. Embeddings between $\langle$bos$\rangle$ and $\langle$pad$\rangle$ are inserted into $\mathcal{W}_p$ to create a new sequence, denoted as $\overline{\mathcal{W}}_i$. }
We use \red{$\mathcal{\overline{W}}_{\text{p}}$} to obtain \red{$K_{\text{p}}$} and $M^{t}$ as in Eq. \ref{eq:attn_map}, assuming that the full-prompt key is able to generate layouts without missing objects. We obtain a set of values from $\mathbb{W}$ and multiply each with $M^t$ to achieve a conjunction of $k$ NPs in $\mathcal{C}$:
\begin{equation}
    \mathbb{V} = [\red{f_{V}(\mathcal{W}_{\text{p}}), f_{V}(\mathcal{\overline{W}}_1),\ldots, f_{V}(\mathcal{\overline{W}}_k)}]=[\red{V_{\text{p}}}, V_1,\dots, V_k].
\end{equation}
\begin{equation} \label{eq:mv}
    O^{t} = \frac{1}{(k+1)}\sum_{i}(M^{t}V_{i}), i=\text{p},1,2,\ldots,k.
\end{equation}

\begin{algorithm}[t]
\caption{StructureDiffusion Guidance. 
}
\label{algo:algo}
\begin{algorithmic}[1]

\REQUIRE ~~\\
\textbf{Input:} Prompt $\mathcal{P}$, Parser $\xi$, decoder $\psi$, trained diffusion model $\phi$.\\
\textbf{Output:} Generated image $x$.\\

\STATE Retrieve concept set $\mathcal{C}=[c_1,\ldots,c_k]$ by traversing $\xi(\mathcal{P})$;\\

\STATE $\mathcal{W}_{\text{p}} \leftarrow \text{CLIP}_{\text{text}}(\mathcal{P})$, \hspace{1pt} $\mathcal{W}_{i} \leftarrow \text{CLIP}_{\text{text}}(c_{i}); \hfill  i=1,\ldots,k$ \\

\FOR{$t=T,T-1,\ldots,1$}
\FOR{each cross attention layer in $\phi$}
\STATE Obtain previous layer's output $\mathcal{X}^t$.
\STATE $Q^t \leftarrow f_{Q}(\mathcal{X}^{t})$, \hspace{1pt} $K_{\text{p}} \leftarrow f_{K}(\mathcal{W}_{\text{p}})$, \hspace{1pt} $V_{i} \leftarrow f_{V}(\mathcal{\overline{W}}_i); \hfill i=\text{p}, 1,\ldots, k$\\

\STATE Obtain attention maps $M^{t}$  from $Q^t, K_{\text{p}}$; \hfill\COMMENT{Eq. \ref{eq:attn_map}} \\

\STATE Obtain ${O}^t$ from $M^{t}, \{V_i\}$, and feed to following layers; \hfill\COMMENT{Eq. \ref{eq:mv}} \\

% \STATE Feed ${O}^t$ to following layers and output $z^{t-1}$
\ENDFOR
\ENDFOR
\STATE Feed $z^0$ to decoder $\psi(\cdot)$ to generate x.
\end{algorithmic}
\end{algorithm}

\red{Compared to using $f_V(\mathcal{W}_p)$ only, Eq. \ref{eq:mv} does not modify the image layout or composition since $M^t$ is still calculated from $Q^t, K_p$. Empirically, we justify the claim by a series of visualizations of} \red{$M_t$} (\red{see Appendix \ref{sec:app_vis}}). However, Stable Diffusion tends to omit objects in generated images (Fig. \ref{fig:teaser}), especially for concept conjunctions that connect two objects with the word ``and''. We devise a variant of our method that computes a set of attention maps $\mathbb{M}=\{M^{t}_{p}, M^{t}_{1}, \ldots \}$ from $\mathcal{C}$ and multiply them to $\mathbb{V}$: 
\begin{equation}
    \red{\mathbb{K} = \{f_{K}(\mathcal{W}_i)\}, \; \mathbb{M}^{t} = \{f_{M}(Q^t, K_i)\}}, \; i=\red{\text{p}},1,2,\ldots,k.
\end{equation}
\begin{equation} \label{eq:mv2}
    \red{O^{t} = \frac{1}{(k+1)}\sum_{i}(M_{i}^{t}V_{k})}, i=\red{\text{p}},1,2,\ldots,k.
\end{equation}
$O^{t}$ is the output of a certain cross-attention layer and the input into downstream layers to generate final image $x$. 
Our algorithm can be summarized as \ref{algo:algo}, which requires no training or additional data.

\section{Experiment} \label{sec:exp}

\subsection{Experiment Settings}
\paragraph{Datasets} 
To address attribute binding and compositional generation, we propose a new benchmark, \textbf{A}ttribute \textbf{B}inding \textbf{C}ontrast set (ABC-6K). It consists of natural prompts from MSCOCO where each contains at least two color words modifying different objects. We also switch the position of two color words to create a contrast caption \citep{gardner2020evaluating}. We end up with 6.4K captions or 3.2K contrastive pairs. In addition to natural compositional prompts, we challenge our method with less detailed prompts that conjunct two concepts together. These prompts follow the sentence pattern of ``a red apple and a yellow banana'' and conjunct two objects with their attribute descriptions. We refer to this set of prompts as \textbf{C}oncept \textbf{C}onjunction 500 (CC-500). We also evaluate our method on 10K randomly sampled captions from MSCOCO \citep{lin2014microsoft}. We show that our method generalizes beyond attribute binding and introduces no quality degradation for general prompts.

\paragraph{Evaluation Metrics}
We mainly rely on human evaluations for compositional prompts and concept conjunction (ABC-6K \& CC-500). We ask annotators to compare two generated images, from Stable Diffusion and our method respectively, and indicate which image demonstrates better image-text alignment or image fidelity. \red{For image fidelity, we ask the annotators ``Regardless of the text, which image is more realistic and natural?''.} \red{We also investigate an automatic evaluation metric for image compositions, i.e., using a SOTA phrase grounding model GLIP \citep{glip} to match phrase-object pairs.} 
\red{As for system-level evaluation}, we follow previous work to utilize Inception Score (IS) \citep{is}, Fre\'{c}het Inception Distance (FID) \citep{fid} and CLIP R-precision (R-prec.) \citep{park2021benchmark}. IS and FID mainly measure the image bank's systematic quality and diversity, while R-prec measures image-level alignment.

\subsection{Compositional Prompts}

\begin{table*}[t]
\centering
\resizebox{\textwidth}{!}{%    
\begin{tabular}{l c ccc ccc}
\toprule
\multirow{2}{*}{ \textbf{Benchmark}} & \multirow{2}{*}{\makecell{StructureDiffusion (ours) \\ \textbf{v.s.}}} & \multicolumn{3}{c}{\textbf{Alignment}} & \multicolumn{3}{c}{\textbf{Fidelity}} \\
\cmidrule(lr){3-5}\cmidrule(lr){6-8}
 &   & Win ($\uparrow$) & \red{Lose ($\downarrow$)} & \red{Tie} & Win ($\uparrow$) & \red{Lose ($\downarrow$)} & \red{Tie} \\
    \midrule
    \multirow{1}{*}{\textbf{ABC-6K}} & Stable Diffusion & \textbf{42.2} & 35.6 & 22.2 & \textbf{48.3} & 39.1 & 12.6 \\
    \midrule
    \multirow{2}{*}{\textbf{CC-500}} & Stable Diffusion & \textbf{31.8} & 27.7 & 38.9 & \textbf{37.8} & 30.6 & 31.6 \\
                                     & Composable Diffusion & \textbf{46.5} & 30.1 & 22.8 & \textbf{61.4} & 19.8 & 18.8 \\
    \bottomrule
\end{tabular}
}
\caption{Percentage of generated images of StructureDiffusion that are better than (win), tied with, or worse than (lose) the compared model in terms of text-image alignment and image fidelity. We filtered out 20\% most similar image pairs for comparison (See Sec. \ref{sec:limits}). Composable Diffusion cannot be applied to ABC-6K as those prompts may not contain explicit ``and'' words that separate concepts. 
}
\label{tab:human}
\end{table*}

\begin{figure*}[t]
    \centering
    \includegraphics[width=0.95\textwidth]{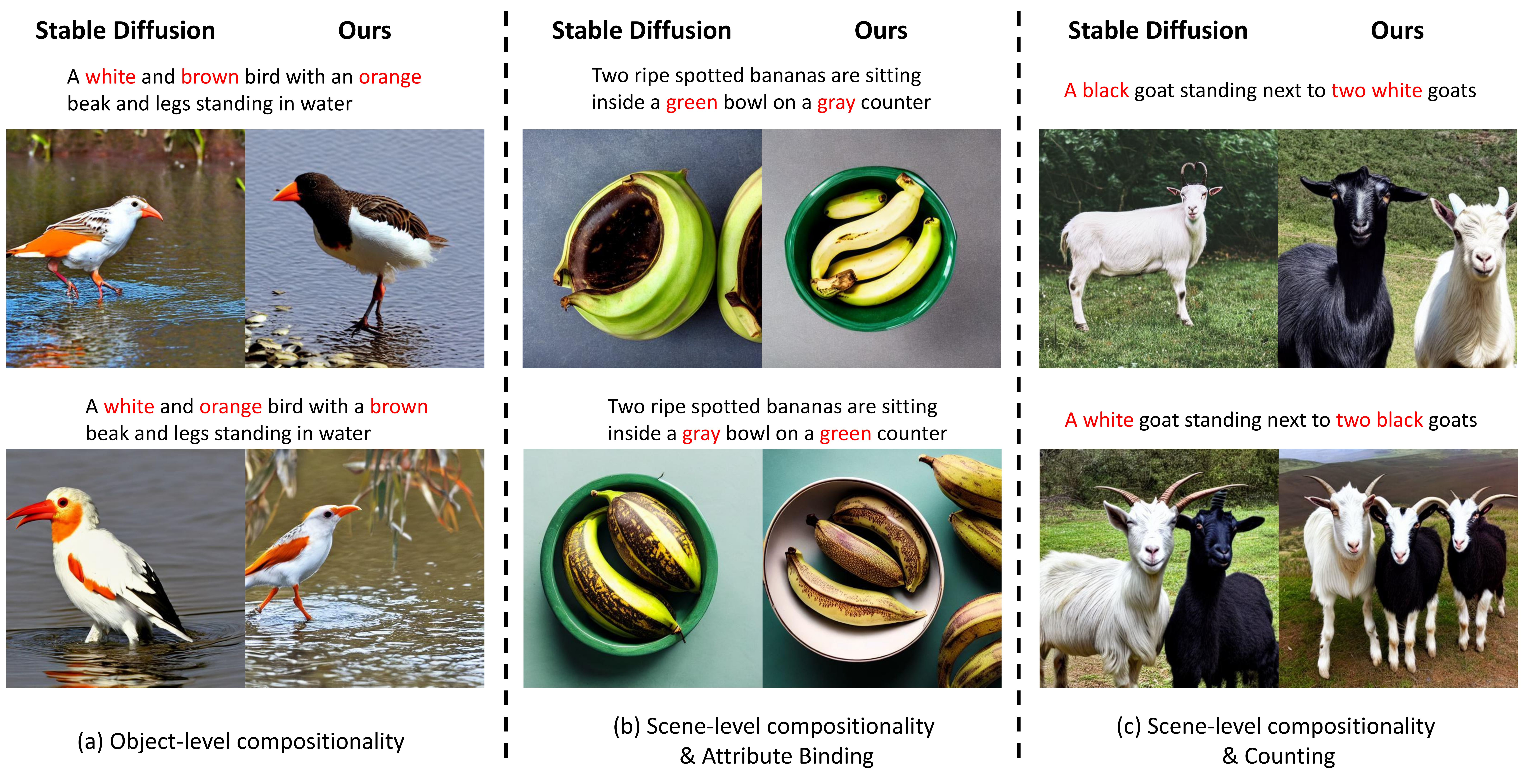}
    \caption{Qualitative results on ABC-6K. Our method improves both object-level and scene-level compositionality.
    }
    \label{fig:abc}
\end{figure*}

Here we show the quantitative and qualitative evaluation results on ABC-6K. We observe that our method sometimes generates very similar images to Stable Diffusion. Hence, we first generate two images per prompt for our method and Stable Diffusion, involving around 12K image pairs to compare. Then, we filter out 20\% of the most similar pairs and then randomly sampled 1500 pairs for human evaluations. As shown in Table \ref{tab:human}, annotators indicate around a 42\% chance of our method winning the comparison, 7\% higher than losing the comparison. There is still a 22\% of chance that our images are \red{tied} with images from Stable Diffusion. 

We show qualitative examples characterizing three different perspectives in Fig. \ref{fig:abc}. \red{Our method fills in the correct color for different parts of an object or different objects, as shown in the first two examples. The third example demonstrates that our method can mitigate the issue of ``missing objects''. Among the 42\% winning cases, there are 31\% for ``fewer missing objects'', 14.1\% for ``better-matched colors'', and 54.8\% for ``other attributes or details'' as indicated by annotators. The results certify that the improvement goes beyond colors to component completeness and fine-grained details. More qualitative examples characterizing all three aspects can be found in Fig. \ref{fig:abc_appendix} in the Appendix.}

\subsection{Concept Conjunction}

\begin{table*}[t]
\centering
\resizebox{\textwidth}{!}{%    
\begin{tabular}{l ccc cc c}
\toprule
    & \multicolumn{6}{c}{\textbf{CC-500} (Prompt format: ``\textit{a} [\textit{colorA}] [\textit{objectA}] and \textit{a} [\textit{colorB}] [\textit{objectB}]'' ) } \\
    \cmidrule(lr){2-7}
    & \multicolumn{3}{c}{Human Annotations} & \multicolumn{2}{c}{GLIP} & \multirow{2}{*}[-1em]{\makecell{Human-GLIP \\Consistency}} \\
    \cmidrule(lr){2-4}\cmidrule(lr){5-6}
    Methods  & \makecell{Zero/One\\ obj. ($\downarrow$)} & Two obj. & \makecell{Two obj. w/ \\correct colors} & \makecell{Zero/One\\ obj. ($\downarrow$)} & Two obj. \\
    \midrule
    \textbf{Stable Diffusion} & \red{65.5} & \red{34.5} & \red{19.2} & \red{69.0} & \red{31.0} & \red{46.4} \\
    \midrule
    \textbf{Composable Diffusion} &  \red{69.7} & \red{30.3} & \red{20.6} & \red{74.2} & \red{25.8} & \red{48.9}\\
    \midrule
    \textbf{StructureDiffusion (Ours)} & \red{\textbf{62.0}} & \red{\textbf{38.0}} & \red{\textbf{22.7}} & \red{\textbf{68.8}} & \red{\textbf{31.2}} & \red{47.6}\\
    \bottomrule
\end{tabular}
}
\caption{\red{Fine-grained human and automatic evaluation results on CC-500. Recall that each prompt is a conjunction of two different objects with different colors. ``Zero/One obj.'' means that the model fails to generate all desired objects in the image. ``Human-GLIP consistency'' reflects the percentage of images where human annotations align with GLIP detection results.}
}
\label{tab:counting}
\end{table*}

\begin{figure*}[t]
    \centering
    \includegraphics[width=0.95\textwidth]{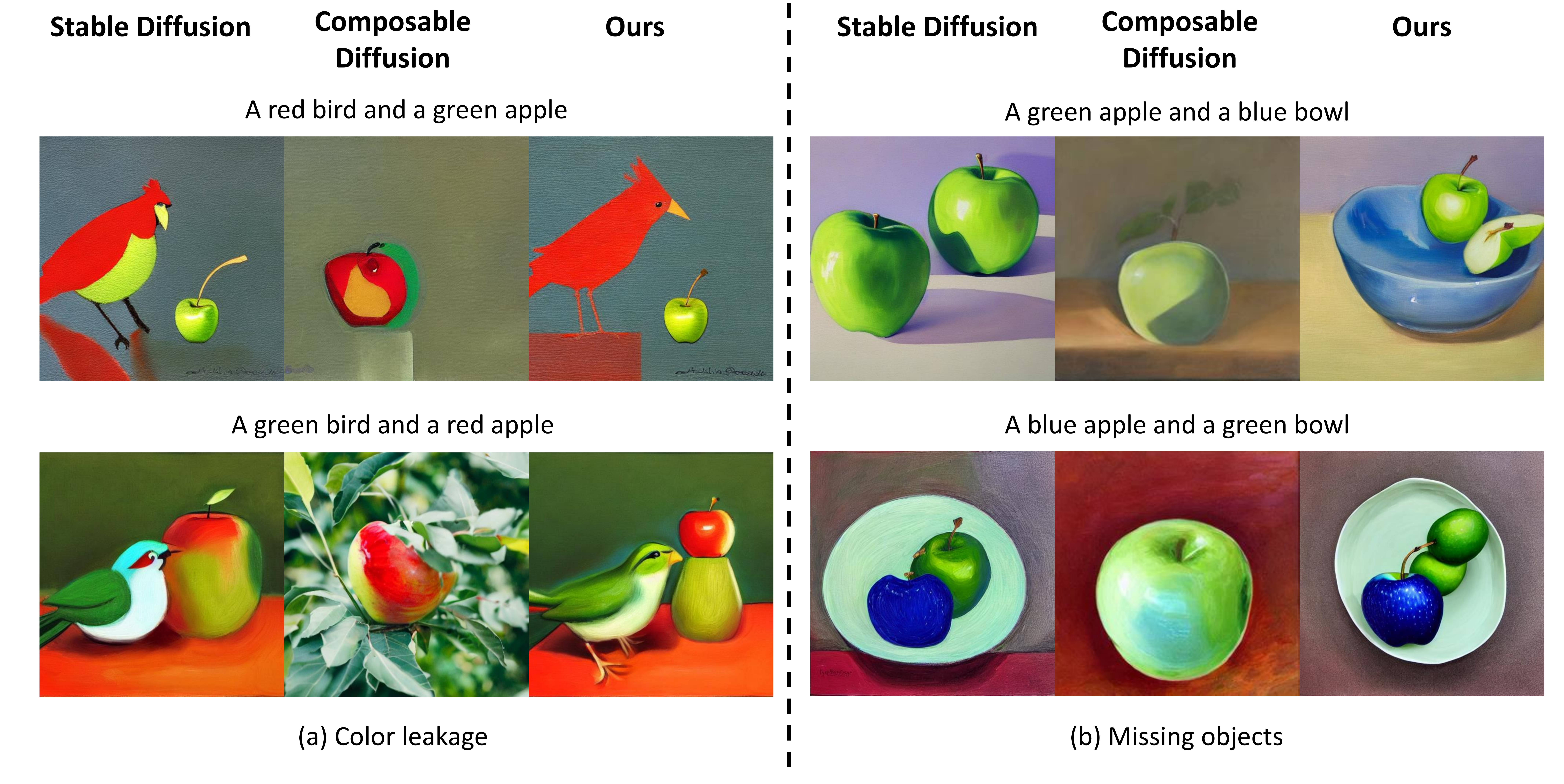}
    \caption{Qualitative results on CC-500 prompts that emphasize two aspects. (a) Color leakage: our method prevents the green color from invading the bird or apple. (b) Missing objects: our method completes the ``blue bowl'' and improves the quality of the ``blue apple''.
    }
    \label{fig:and}
\end{figure*}

Here we address challenging concept conjunction prompts and evaluate our method on CC-500. Apart from Stable Diffusion, we also compare to Composable Diffusion \citep{liu2022compositional} implemented on top of Stable Diffusion. For Composable Diffusion, we separate the prompts into text segments by the keyword ``and'' and feed each span into an independent diffusion process. We generate three images per prompt and use all images for human evaluation for Stable Diffusion. We randomly sampled 600 images for comparison to Composable Diffusion.

As shown in Table \ref{tab:human}, our method outperforms Stable Diffusion by around 4.1\% and Composable Diffusion by 16.4\% in terms of image-text alignment. We also observe that our method enhances some fine-grained details in the generated images, leading to a 7.2\% improvement in image fidelity when compared with Stable Diffusion. We observe that images from composable diffusion can be oversaturated with unnatural visual textures and layouts, which could be the reason for StructureDiffusion to have high win rate in image fidelity. \red{As shown in Fig. \ref{fig:and} and Fig. \ref{fig:cb_appendix}. Our approach prevents color bleeding (left), missing objects (right) and strengthens details (right).}

\red{To further quantify the text-image alignment, we consider both human annotations and automatic evaluations. For each object mentioned in the prompt, we ask annotators whether the object exists in the image and whether it is in the correct color. We also apply a state-of-the-art detection model GLIP \citep{glip} to ground each ``\textit{a} [\textit{color}] [\textit{object}]'' phrase into bounding boxes. We report the percentage of images that contain incomplete objects / complete objects / complete objects with correct colors in Table \ref{tab:counting}. StructureDiffusion improves the compositionality by 3.5\% based on human annotations while only 0.2\% based on GLIP. We discover that humans disagree with GLIP for more than 50\% of the images, as entailed by the low consistency rate. Previous work also suggests the deficiency of large pre-trained models in compositional understanding \citep{thrush2022winoground}.}

\subsection{Other prompts}

We show that our StructureDiffusion maintain the overall image quality and diversity on general prompts. We follow the standard evaluation process and generate 10,000 images from randomly sampled MSCOCO captions. Stable Diffusion obtains 39.9 IS, 18.0 FID and 72.2 R-Precision. Our method achieves 40.9 IS, 17.9 FID and 72.3 R-Precision. StructureDiffusion maintains the image fidelity and diversity as indicated in the comparable IS/FID/R-Prec scores.

\begin{figure*}[t]
    \centering
    \includegraphics[width=\textwidth]{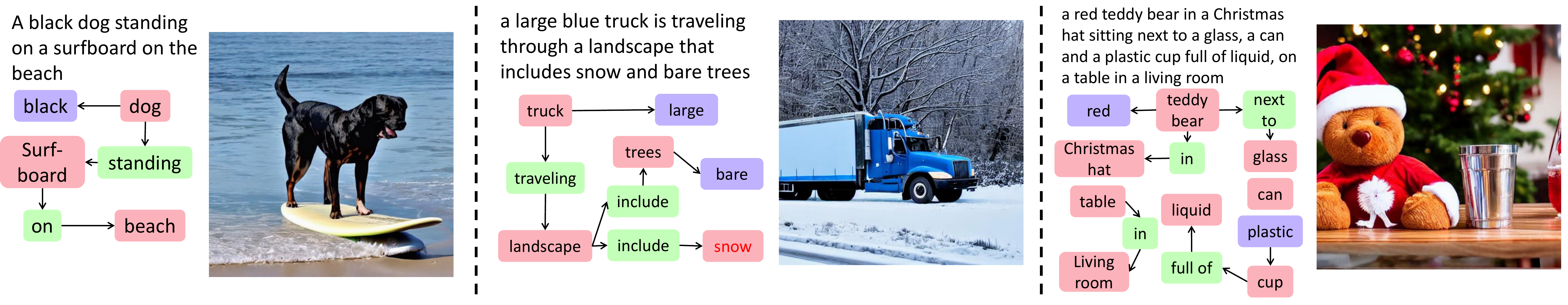}
    \caption{\red{Qualitative results of using scene graph parser to generate structured representations.}
    }
    \label{fig:scene_graph}
\end{figure*}

\subsection{Scene Graph Input}
We show that our method is not limited to constituency parsing but can also be extended to other structured representations, such as scene graphs. As shown in Fig. \ref{fig:scene_graph}, we first adopt the scene graph parser \citep{wu2019unified} and obtain a graph like the ones next to each image from the input prompt. The parser returns basic entities and their relations in between. We extract text spans of basic entities with their attributes attached and text spans that include two related entities. \red{We provide examples in Appendix \ref{tab:parsers} and make comparison to the constituency parser. Similarly, we encode these spans separately and re-align each with the entire prompt encoding sequence. On MS-COCO, the scene graph parser setting maintains the image quality with 39.2 IS, 17.9 FID, and 72.0 R-Precision. When compared to Stable Diffusion on ABC-6K, the scene graph parser achieves 34.2\%-32.9\%-32.9\% Win-Lose-Tie in image-text alignment and 34.5\%-32.5\%-33.0\% Win-Lose-Tie in image fidelity. As for CC-500, the scene graph parser leads to the same output images due to the same text spans. We refer to Table \ref{tab:parsers} and Fig. \ref{fig:sg} for more results and comparison.}

\section{Ablation Study}

\subsection{\red{Re-aligning Sequence}}
\red{In Section \ref{sec:method}, we describe a method to realign the encoding of a text span back into the sequence of the full prompt. Since the noun-phrase text spans are shorter than the full sequence, re-alignment ensures that each token's value vector corresponds to the correct attention map. On the other hand, naively expanding the span to the length of the full sequence degrades the image quality by $\sim$2 IS / FID (37.5 IS, 19.8 FID) compared to images with re-alignment or Stable Diffusion.}

\begin{figure*}[t]
    \centering
    \includegraphics[width=0.95\textwidth]{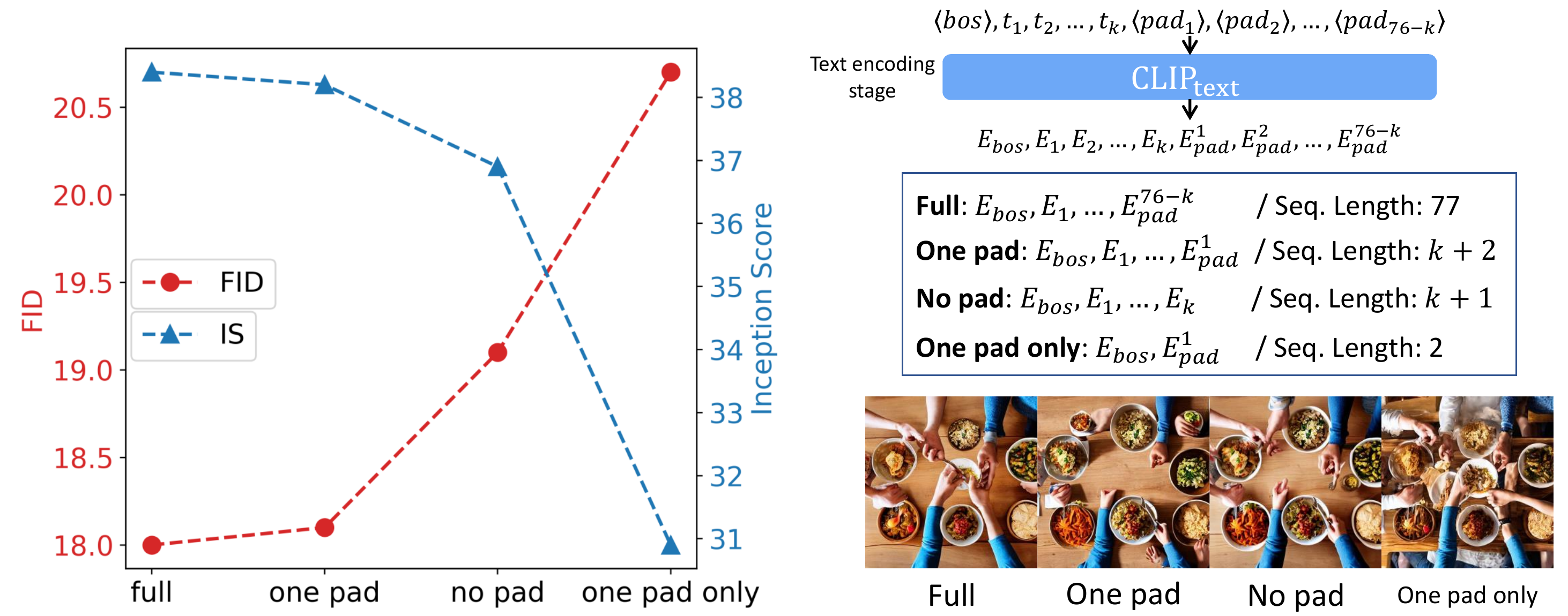}
    \caption{Ablation study on the text sequence embeddings. We find that the padding embeddings are fully contextualized, representing the prompt's high-level semantics. However, not all padding tokens are necessary to maintain a high-fidelity output from Stable Diffusion. 
    }
    \label{fig:efficient}
\end{figure*}

\subsection{Contextualized Text Embeddings}
One limitation brought by our StructureDiffusion is that the cross-attention computation costs increase by the number of noun phrases. Yet we noticed that most of the attention maps are computed from padding embeddings, as Stable Diffusion adopts CLIP text encoders and automatically pads the sequence to 77 tokens. We conjecture that not all padding tokens are necessary for generating high-quality images. As is shown in Fig. \ref{fig:efficient}, we study four different patterns of token embeddings. We discover that leaving the nearest padding embeddings maintains a similar IS / FID score as the full sequence. Further removing this padding embedding results in apparent degradation. While only using the nearest padding embedding results in the worst image quality, we find that the high-level image layout and semantics are preserved (see bottom right of Fig. \ref{fig:efficient}). This phenomenon indicates that the padding embeddings are fully contextualized with the full prompt semantics. This also justifies our re-alignment operation that preserves padding embeddings of the main sequence $\mathcal{W}_{\text{full}}$.

\section{Related Work} \label{sec:related}

\paragraph{Text-to-Image Synthesis}
The diffusion model is an emerging type of model that generate high-quality images with a much more stable training process \citep{song2019generative, ddpm}.  \citet{rombach2022high} proposes to encode an image with an autoencoder and then leverage a diffusion model to generate continuous feature maps in the latent space. Stable Diffusion \citet{rombach2022high} adopts similar architecture but is trained on large-scale image-text datasets with fixed CLIP text encoder. Imagen \citep{imagen} addresses the importance of language understanding by using a frozen T5 encoder \citep{t5}, a dedicated large language model. We mainly focus on diffusion models and conduct our experiments on Stable Diffusion \citep{rombach2022high}, the SOTA open-sourced T2I model.

\paragraph{Compositional Generation}
The compositional or controllable generation has been an essential direction for T2I models to understand and disentangle basic concepts in the generation process. As text inputs are relatively weak conditions, previous work leverage layout or scene graph to enhance compositionality \citep{johnson2018image, hong2018inferring, yang2022modeling, gafni2022make}. More recently, \citet{liu2022compositional} proposes an approach where the concept conjunctions are achieved by adding estimated scores from a parallel set of diffusion processes. In contrast, our method can be directly merged into the cross-attention layers with much less computational overhead.

\paragraph{Diffusion Guidance} 
\citet{ho2022classifier} develops classifier-free guidance where a single diffusion model is jointly trained under conditional and unconditional inputs. Most large-scale SOTA models, including autoregressive ones, adopt this technique for flexible and improved conditional synthesis results \citep{rombach2022high, dalle2, gafni2022make, parti, imagen}. \citet{hertz2022prompt} discovers unique properties of cross attention maps on Imagen \citep{imagen} and achieves structure-preserving image editing by manipulating these maps. We observe similar properties in Stable Diffusion \citep{rombach2022high} but propose a different algorithm for fine-grained, compositional text-to-image generation. 

\section{Conclusion} \label{sec:conclusion}
In this work, we propose a training-free method for compositional text-to-image generation. First, we observe that existing large-scale T2I diffusion models can still struggle in compositional image synthesis. We address this challenge by explicitly focusing on binding objects with the correct attributes. Second, we propose structured diffusion guidance incorporating language structures into the cross-attention layers. We propose two simple techniques to align the structured encoding with the attention maps. Using our structured guidance on Stable Diffusion, attributes can be bound more accurately while maintaining the overall image quality and diversity. In addition, we justify our approach by conducting an in-depth analysis of the frozen language encoder and attention maps. Future work may explore explicit approaches to generate plausible image layouts without missing components. We hope that our approach accelerates the development of interpretable and efficient methods for diffusion-based text-to-image models. 
% \section{Limitations \& Future Work}

\section*{Acknowledgement}
We would like to thank the Robert N. Noyce Trust for their generous gift to the University of California via the Noyce Initiative. The work was also partially funded by an unrestricted gift from Google and by the National Science Foundation award \#2048122. The writers’ opinions and conclusions in this publication are their own and should not be construed as representing the sponsors’ official policy, expressed or inferred.

\section*{Reproducibility Statement}
We release our core codebase containing the methodology implementation, settings, benchmarks containing compositional prompts under supplementary materials.

\section*{Ethical Statement}
As for the data collection and verification, we use the Amazon Mechanical Turk platform and form the comparison task as batches of HITs. We select workers from English-speaking countries, including the US, CA, UK, AU, and NZ, since the task require understanding the English input prompt. Each HIT takes around 15-30 seconds on average to accomplish, and we pay each submitted HIT with 0.15 US dollars, resulting in an hourly payment of 18 US dollars.  

% \subsubsection*{Acknowledgments}
% Use unnumbered third level headings for the acknowledgments. All
% acknowledgments, including those to funding agencies, go at the end of the paper.

\bibliography{iclr2023_conference}

\begin{thebibliography}{46}
\providecommand{\natexlab}[1]{#1}
\providecommand{\url}[1]{\texttt{#1}}
\expandafter\ifx\csname urlstyle\endcsname\relax
  \providecommand{\doi}[1]{doi: #1}\else
  \providecommand{\doi}{doi: \begingroup \urlstyle{rm}\Url}\fi

\bibitem[Dhariwal \& Nichol(2021)Dhariwal and Nichol]{dhariwal2021diffusion}
Prafulla Dhariwal and Alexander Nichol.
\newblock Diffusion models beat gans on image synthesis.
\newblock \emph{Advances in Neural Information Processing Systems},
  34:\penalty0 8780--8794, 2021.

\bibitem[Ding et~al.(2022)Ding, Zheng, Hong, and Tang]{cogview2}
Ming Ding, Wendi Zheng, Wenyi Hong, and Jie Tang.
\newblock Cogview2: Faster and better text-to-image generation via hierarchical
  transformers.
\newblock \emph{arXiv preprint arXiv:2204.14217}, 2022.

\bibitem[El-Nouby et~al.(2019)El-Nouby, Sharma, Schulz, Hjelm, Asri, Kahou,
  Bengio, and W.Taylor]{el-nouby2019ilbie}
Alaaeldin El-Nouby, Shikhar Sharma, Hannes Schulz, Devon Hjelm, Layla~El Asri,
  Samira~Ebrahimi Kahou, Yoshua Bengio, and Graham W.Taylor.
\newblock {Tell, Draw, and Repeat: Generating and Modifying Images Based on
  Continual Linguistic Instruction}.
\newblock In \emph{ICCV}, 2019.

\bibitem[Fu et~al.(2020)Fu, Wang, Grafton, Eckstein, and Wang]{fu2020sscr}
Tsu-Jui Fu, Xin~Eric Wang, Scott Grafton, Miguel Eckstein, and William~Yang
  Wang.
\newblock {SSCR: Iterative Language-Based Image Editing via Self-Supervised
  Counterfactual Reasoning}.
\newblock In \emph{EMNLP}, 2020.

\bibitem[Gafni et~al.(2022)Gafni, Polyak, Ashual, Sheynin, Parikh, and
  Taigman]{gafni2022make}
Oran Gafni, Adam Polyak, Oron Ashual, Shelly Sheynin, Devi Parikh, and Yaniv
  Taigman.
\newblock Make-a-scene: Scene-based text-to-image generation with human priors.
\newblock \emph{arXiv preprint arXiv:2203.13131}, 2022.

\bibitem[Gardner et~al.(2020)Gardner, Artzi, Basmov, Berant, Bogin, Chen,
  Dasigi, Dua, Elazar, Gottumukkala, et~al.]{gardner2020evaluating}
Matt Gardner, Yoav Artzi, Victoria Basmov, Jonathan Berant, Ben Bogin, Sihao
  Chen, Pradeep Dasigi, Dheeru Dua, Yanai Elazar, Ananth Gottumukkala, et~al.
\newblock Evaluating models’ local decision boundaries via contrast sets.
\newblock \emph{Findings of Empirical Methods in Natural Language Processing},
  2020.

\bibitem[Gu et~al.(2022{\natexlab{a}})Gu, Chen, Bao, Wen, Zhang, Chen, Yuan,
  and Guo]{gu2022vector}
Shuyang Gu, Dong Chen, Jianmin Bao, Fang Wen, Bo~Zhang, Dongdong Chen, Lu~Yuan,
  and Baining Guo.
\newblock Vector quantized diffusion model for text-to-image synthesis.
\newblock In \emph{Proceedings of the IEEE/CVF Conference on Computer Vision
  and Pattern Recognition}, pp.\  10696--10706, 2022{\natexlab{a}}.

\bibitem[Gu et~al.(2022{\natexlab{b}})Gu, Chen, Bao, Wen, Zhang, Chen, Yuan,
  and Guo]{vqtransformer}
Shuyang Gu, Dong Chen, Jianmin Bao, Fang Wen, Bo~Zhang, Dongdong Chen, Lu~Yuan,
  and Baining Guo.
\newblock Vector quantized diffusion model for text-to-image synthesis.
\newblock In \emph{Proceedings of the IEEE/CVF Conference on Computer Vision
  and Pattern Recognition}, pp.\  10696--10706, 2022{\natexlab{b}}.

\bibitem[Hertz et~al.(2022)Hertz, Mokady, Tenenbaum, Aberman, Pritch, and
  Cohen-Or]{hertz2022prompt}
Amir Hertz, Ron Mokady, Jay Tenenbaum, Kfir Aberman, Yael Pritch, and Daniel
  Cohen-Or.
\newblock Prompt-to-prompt image editing with cross attention control.
\newblock \emph{arXiv preprint arXiv:2208.01626}, 2022.

\bibitem[Heusel et~al.(2017)Heusel, Ramsauer, Unterthiner, Nessler, and
  Hochreiter]{fid}
Martin Heusel, Hubert Ramsauer, Thomas Unterthiner, Bernhard Nessler, and Sepp
  Hochreiter.
\newblock Gans trained by a two time-scale update rule converge to a local nash
  equilibrium.
\newblock \emph{Advances in neural information processing systems}, 30, 2017.

\bibitem[Ho \& Salimans(2022)Ho and Salimans]{ho2022classifier}
Jonathan Ho and Tim Salimans.
\newblock Classifier-free diffusion guidance.
\newblock \emph{arXiv preprint arXiv:2207.12598}, 2022.

\bibitem[Ho et~al.(2020)Ho, Jain, and Abbeel]{ddpm}
Jonathan Ho, Ajay Jain, and Pieter Abbeel.
\newblock Denoising diffusion probabilistic models.
\newblock \emph{Advances in Neural Information Processing Systems},
  33:\penalty0 6840--6851, 2020.

\bibitem[Hong et~al.(2018)Hong, Yang, Choi, and Lee]{hong2018inferring}
Seunghoon Hong, Dingdong Yang, Jongwook Choi, and Honglak Lee.
\newblock Inferring semantic layout for hierarchical text-to-image synthesis.
\newblock In \emph{Proceedings of the IEEE conference on computer vision and
  pattern recognition}, pp.\  7986--7994, 2018.

\bibitem[Johnson et~al.(2018)Johnson, Gupta, and Fei-Fei]{johnson2018image}
Justin Johnson, Agrim Gupta, and Li~Fei-Fei.
\newblock Image generation from scene graphs.
\newblock In \emph{Proceedings of the IEEE conference on computer vision and
  pattern recognition}, pp.\  1219--1228, 2018.

\bibitem[Lee et~al.(2022)Lee, Kim, Kim, Cho, and Han]{lee2022autoregressive}
Doyup Lee, Chiheon Kim, Saehoon Kim, Minsu Cho, and Wook-Shin Han.
\newblock Autoregressive image generation using residual quantization.
\newblock In \emph{Proceedings of the IEEE/CVF Conference on Computer Vision
  and Pattern Recognition}, pp.\  11523--11532, 2022.

\bibitem[Li et~al.(2019)Li, Qi, Lukasiewicz, and Torr]{controlgan}
Bowen Li, Xiaojuan Qi, Thomas Lukasiewicz, and Philip Torr.
\newblock Controllable text-to-image generation.
\newblock \emph{Advances in Neural Information Processing Systems}, 32, 2019.

\bibitem[Li et~al.(2022)Li, Zhang, Zhang, Yang, Li, Zhong, Wang, Yuan, Zhang,
  Hwang, et~al.]{glip}
Liunian~Harold Li, Pengchuan Zhang, Haotian Zhang, Jianwei Yang, Chunyuan Li,
  Yiwu Zhong, Lijuan Wang, Lu~Yuan, Lei Zhang, Jenq-Neng Hwang, et~al.
\newblock Grounded language-image pre-training.
\newblock In \emph{Proceedings of the IEEE/CVF Conference on Computer Vision
  and Pattern Recognition}, pp.\  10965--10975, 2022.

\bibitem[Lin et~al.(2014)Lin, Maire, Belongie, Hays, Perona, Ramanan,
  Doll{\'a}r, and Zitnick]{lin2014microsoft}
Tsung-Yi Lin, Michael Maire, Serge Belongie, James Hays, Pietro Perona, Deva
  Ramanan, Piotr Doll{\'a}r, and C~Lawrence Zitnick.
\newblock Microsoft coco: Common objects in context.
\newblock In \emph{European conference on computer vision}, pp.\  740--755.
  Springer, 2014.

\bibitem[Liu et~al.(2021{\natexlab{a}})Liu, Ren, Lin, and Zhao]{liu2021pseudo}
Luping Liu, Yi~Ren, Zhijie Lin, and Zhou Zhao.
\newblock Pseudo numerical methods for diffusion models on manifolds.
\newblock In \emph{International Conference on Learning Representations},
  2021{\natexlab{a}}.

\bibitem[Liu et~al.(2022)Liu, Li, Du, Torralba, and
  Tenenbaum]{liu2022compositional}
Nan Liu, Shuang Li, Yilun Du, Antonio Torralba, and Joshua~B Tenenbaum.
\newblock Compositional visual generation with composable diffusion models.
\newblock \emph{arXiv preprint arXiv:2206.01714}, 2022.

\bibitem[Liu et~al.(2021{\natexlab{b}})Liu, Park, Azadi, Zhang, Chopikyan, Hu,
  Shi, Rohrbach, and Darrell]{liu2021more}
Xihui Liu, Dong~Huk Park, Samaneh Azadi, Gong Zhang, Arman Chopikyan, Yuxiao
  Hu, Humphrey Shi, Anna Rohrbach, and Trevor Darrell.
\newblock More control for free! image synthesis with semantic diffusion
  guidance.
\newblock \emph{arXiv preprint arXiv:2112.05744}, 2021{\natexlab{b}}.

\bibitem[Lou et~al.(2022)Lou, Han, Lin, and Zheng]{Lou_2022_CVPR}
Chao Lou, Wenjuan Han, Yuhuan Lin, and Zilong Zheng.
\newblock Unsupervised vision-language parsing: Seamlessly bridging visual
  scene graphs with language structures via dependency relationships.
\newblock In \emph{Proceedings of the IEEE/CVF Conference on Computer Vision
  and Pattern Recognition (CVPR)}, pp.\  15607--15616, June 2022.

\bibitem[Nichol et~al.(2021)Nichol, Dhariwal, Ramesh, Shyam, Mishkin, McGrew,
  Sutskever, and Chen]{glide}
Alex Nichol, Prafulla Dhariwal, Aditya Ramesh, Pranav Shyam, Pamela Mishkin,
  Bob McGrew, Ilya Sutskever, and Mark Chen.
\newblock Glide: Towards photorealistic image generation and editing with
  text-guided diffusion models.
\newblock \emph{ICML}, 2021.

\bibitem[Park et~al.(2021)Park, Azadi, Liu, Darrell, and
  Rohrbach]{park2021benchmark}
Dong~Huk Park, Samaneh Azadi, Xihui Liu, Trevor Darrell, and Anna Rohrbach.
\newblock Benchmark for compositional text-to-image synthesis.
\newblock In \emph{Thirty-fifth Conference on Neural Information Processing
  Systems Datasets and Benchmarks Track (Round 1)}, 2021.

\bibitem[Qi et~al.(2020)Qi, Zhang, Zhang, Bolton, and Manning]{qi2020stanza}
Peng Qi, Yuhao Zhang, Yuhui Zhang, Jason Bolton, and Christopher~D. Manning.
\newblock Stanza: A {Python} natural language processing toolkit for many human
  languages.
\newblock In \emph{Proceedings of the 58th Annual Meeting of the Association
  for Computational Linguistics: System Demonstrations}, 2020.

\bibitem[Radford et~al.(2021)Radford, Kim, Hallacy, Ramesh, Goh, Agarwal,
  Sastry, Askell, Mishkin, Clark, et~al.]{clip}
Alec Radford, Jong~Wook Kim, Chris Hallacy, Aditya Ramesh, Gabriel Goh,
  Sandhini Agarwal, Girish Sastry, Amanda Askell, Pamela Mishkin, Jack Clark,
  et~al.
\newblock Learning transferable visual models from natural language
  supervision.
\newblock In \emph{International Conference on Machine Learning}, pp.\
  8748--8763. PMLR, 2021.

\bibitem[Raffel et~al.(2020)Raffel, Shazeer, Roberts, Lee, Narang, Matena,
  Zhou, Li, and Liu]{t5}
Colin Raffel, Noam Shazeer, Adam Roberts, Katherine Lee, Sharan Narang, Michael
  Matena, Yanqi Zhou, Wei Li, and Peter~J. Liu.
\newblock Exploring the limits of transfer learning with a unified text-to-text
  transformer.
\newblock \emph{Journal of Machine Learning Research}, 21\penalty0
  (140):\penalty0 1--67, 2020.
\newblock URL \url{http://jmlr.org/papers/v21/20-074.html}.

\bibitem[Ramesh et~al.(2021)Ramesh, Pavlov, Goh, Gray, Voss, Radford, Chen, and
  Sutskever]{ramesh2021zero}
Aditya Ramesh, Mikhail Pavlov, Gabriel Goh, Scott Gray, Chelsea Voss, Alec
  Radford, Mark Chen, and Ilya Sutskever.
\newblock Zero-shot text-to-image generation.
\newblock In \emph{International Conference on Machine Learning}, pp.\
  8821--8831. PMLR, 2021.

\bibitem[Ramesh et~al.(2022)Ramesh, Dhariwal, Nichol, Chu, and Chen]{dalle2}
Aditya Ramesh, Prafulla Dhariwal, Alex Nichol, Casey Chu, and Mark Chen.
\newblock Hierarchical text-conditional image generation with clip latents.
\newblock \emph{arXiv preprint arXiv:2204.06125}, 2022.

\bibitem[Rombach et~al.(2022)Rombach, Blattmann, Lorenz, Esser, and
  Ommer]{rombach2022high}
Robin Rombach, Andreas Blattmann, Dominik Lorenz, Patrick Esser, and Bj{\"o}rn
  Ommer.
\newblock High-resolution image synthesis with latent diffusion models.
\newblock In \emph{Proceedings of the IEEE/CVF Conference on Computer Vision
  and Pattern Recognition}, pp.\  10684--10695, 2022.

\bibitem[Ronneberger et~al.(2015)Ronneberger, Fischer, and Brox]{unet}
Olaf Ronneberger, Philipp Fischer, and Thomas Brox.
\newblock U-net: Convolutional networks for biomedical image segmentation.
\newblock In \emph{International Conference on Medical image computing and
  computer-assisted intervention}, pp.\  234--241. Springer, 2015.

\bibitem[Ruiz et~al.(2022)Ruiz, Li, Jampani, Pritch, Rubinstein, and
  Aberman]{ruiz2022dreambooth}
Nataniel Ruiz, Yuanzhen Li, Varun Jampani, Yael Pritch, Michael Rubinstein, and
  Kfir Aberman.
\newblock Dreambooth: Fine tuning text-to-image diffusion models for
  subject-driven generation.
\newblock \emph{arXiv preprint arXiv:2208.12242}, 2022.

\bibitem[Saharia et~al.(2022)Saharia, Chan, Saxena, Li, Whang, Denton,
  Ghasemipour, Ayan, Mahdavi, Lopes, et~al.]{imagen}
Chitwan Saharia, William Chan, Saurabh Saxena, Lala Li, Jay Whang, Emily
  Denton, Seyed Kamyar~Seyed Ghasemipour, Burcu~Karagol Ayan, S~Sara Mahdavi,
  Rapha~Gontijo Lopes, et~al.
\newblock Photorealistic text-to-image diffusion models with deep language
  understanding.
\newblock \emph{arXiv preprint arXiv:2205.11487}, 2022.

\bibitem[Salimans et~al.(2016)Salimans, Goodfellow, Zaremba, Cheung, Radford,
  and Chen]{is}
Tim Salimans, Ian Goodfellow, Wojciech Zaremba, Vicki Cheung, Alec Radford, and
  Xi~Chen.
\newblock Improved techniques for training gans.
\newblock \emph{Advances in neural information processing systems}, 29, 2016.

\bibitem[Schuster et~al.(2015)Schuster, Krishna, Chang, Fei-Fei, and
  Manning]{schuster2015generating}
Sebastian Schuster, Ranjay Krishna, Angel Chang, Li~Fei-Fei, and Christopher~D
  Manning.
\newblock Generating semantically precise scene graphs from textual
  descriptions for improved image retrieval.
\newblock In \emph{Proceedings of the fourth workshop on vision and language},
  pp.\  70--80, 2015.

\bibitem[Song \& Ermon(2019)Song and Ermon]{song2019generative}
Yang Song and Stefano Ermon.
\newblock Generative modeling by estimating gradients of the data distribution.
\newblock \emph{Advances in Neural Information Processing Systems}, 32, 2019.

\bibitem[Tao et~al.(2022)Tao, Tang, Wu, Jing, Bao, and Xu]{df-gan}
Ming Tao, Hao Tang, Fei Wu, Xiao-Yuan Jing, Bing-Kun Bao, and Changsheng Xu.
\newblock Df-gan: A simple and effective baseline for text-to-image synthesis.
\newblock In \emph{Proceedings of the IEEE/CVF Conference on Computer Vision
  and Pattern Recognition}, pp.\  16515--16525, 2022.

\bibitem[Thrush et~al.(2022)Thrush, Jiang, Bartolo, Singh, Williams, Kiela, and
  Ross]{thrush2022winoground}
Tristan Thrush, Ryan Jiang, Max Bartolo, Amanpreet Singh, Adina Williams, Douwe
  Kiela, and Candace Ross.
\newblock Winoground: Probing vision and language models for visio-linguistic
  compositionality.
\newblock In \emph{Proceedings of the IEEE/CVF Conference on Computer Vision
  and Pattern Recognition}, pp.\  5238--5248, 2022.

\bibitem[Wan et~al.(2021)Wan, Han, Zheng, and Tuytelaars]{wan2021unsupervised}
Bo~Wan, Wenjuan Han, Zilong Zheng, and Tinne Tuytelaars.
\newblock Unsupervised vision-language grammar induction with shared structure
  modeling.
\newblock In \emph{International Conference on Learning Representations}, 2021.

\bibitem[Wu et~al.(2019)Wu, Mao, Zhang, Jiang, Li, Sun, and Ma]{wu2019unified}
Hao Wu, Jiayuan Mao, Yufeng Zhang, Yuning Jiang, Lei Li, Weiwei Sun, and
  Wei-Ying Ma.
\newblock Unified visual-semantic embeddings: Bridging vision and language with
  structured meaning representations.
\newblock In \emph{Proceedings of the IEEE/CVF Conference on Computer Vision
  and Pattern Recognition}, pp.\  6609--6618, 2019.

\bibitem[Yang et~al.(2022)Yang, Liu, Wang, Yang, and Tao]{yang2022modeling}
Zuopeng Yang, Daqing Liu, Chaoyue Wang, Jie Yang, and Dacheng Tao.
\newblock Modeling image composition for complex scene generation.
\newblock In \emph{Proceedings of the IEEE/CVF Conference on Computer Vision
  and Pattern Recognition}, pp.\  7764--7773, 2022.

\bibitem[Yu et~al.(2022)Yu, Xu, Koh, Luong, Baid, Wang, Vasudevan, Ku, Yang,
  Ayan, et~al.]{parti}
Jiahui Yu, Yuanzhong Xu, Jing~Yu Koh, Thang Luong, Gunjan Baid, Zirui Wang,
  Vijay Vasudevan, Alexander Ku, Yinfei Yang, Burcu~Karagol Ayan, et~al.
\newblock Scaling autoregressive models for content-rich text-to-image
  generation.
\newblock \emph{arXiv preprint arXiv:2206.10789}, 2022.

\bibitem[Zhang et~al.(2021)Zhang, Koh, Baldridge, Lee, and
  Yang]{zhang2021cross}
Han Zhang, Jing~Yu Koh, Jason Baldridge, Honglak Lee, and Yinfei Yang.
\newblock Cross-modal contrastive learning for text-to-image generation.
\newblock In \emph{Proceedings of the IEEE/CVF conference on computer vision
  and pattern recognition}, pp.\  833--842, 2021.

\bibitem[Zhong et~al.(2020)Zhong, Wang, Chen, Yu, and
  Li]{zhong2020comprehensive}
Yiwu Zhong, Liwei Wang, Jianshu Chen, Dong Yu, and Yin Li.
\newblock Comprehensive image captioning via scene graph decomposition.
\newblock In \emph{European Conference on Computer Vision}, pp.\  211--229.
  Springer, 2020.

\bibitem[Zhou et~al.(2022)Zhou, Zhang, Chen, Li, Tensmeyer, Yu, Gu, Xu, and
  Sun]{zhou2022towards}
Yufan Zhou, Ruiyi Zhang, Changyou Chen, Chunyuan Li, Chris Tensmeyer, Tong Yu,
  Jiuxiang Gu, Jinhui Xu, and Tong Sun.
\newblock Towards language-free training for text-to-image generation.
\newblock In \emph{Proceedings of the IEEE/CVF Conference on Computer Vision
  and Pattern Recognition}, pp.\  17907--17917, 2022.

\bibitem[Zhu et~al.(2019)Zhu, Pan, Chen, and Yang]{dm-gan}
Minfeng Zhu, Pingbo Pan, Wei Chen, and Yi~Yang.
\newblock Dm-gan: Dynamic memory generative adversarial networks for
  text-to-image synthesis.
\newblock In \emph{Proceedings of the IEEE/CVF conference on computer vision
  and pattern recognition}, pp.\  5802--5810, 2019.

\end{thebibliography}
\bibliographystyle{iclr2023_conference}

\appendix
\section{Related Work}
\paragraph{Text-to-Image Synthesis}
There are mainly three types of models for text-to-image synthesis: GAN-based \citep{df-gan, dm-gan, controlgan, fu2020sscr, el-nouby2019ilbie}, autoregressive \citep{vqtransformer, lee2022autoregressive, cogview2} and diffusion models \citep{liu2021more, glide, ruiz2022dreambooth}. \citet{zhang2021cross} proposes XMC-GAN, a one-stage GAN that employs multiple contrastive losses between image-image, image-text, and region-token pairs. More recently, LAFITE \citep{zhou2022towards} enables language-free training by constructing pseudo image-text feature pairs using CLIP \citep{clip}. As for autoregressive models, DALL-E adopts VQ-VAE to quantize image patches into tokens and then uses a transformer to generate discrete tokens sequentially \citep{ramesh2021zero}. Parti \citep{parti} and Make-A-Scene \citep{gafni2022make} both leverage classifier-free guidance to improve controllability. As for diffusion models, \citet{gu2022vector} concatenates VQ-VAE with the diffusion model and shows that the diffusion process can operate in discrete latent space. DALL-E 2 adopts the CLIP text encoder so that the diffusion process inverts the textual features into images \citep{dalle2}.

\paragraph{Structured Representations for Vision and Language} Inferring shared structures across language and vision has been a long-term pursuit in unifying these modalities \citep{schuster2015generating, johnson2018image, zhong2020comprehensive, Lou_2022_CVPR}. \citet{wu2019unified} utilizes the structure from semantic parsing in a visual-semantic embedding framework to facilitate embedding learning. \citet{wan2021unsupervised} proposes a new task in which the goal is to learn a joint structure between semantic parsing and image regions. To the best of our knowledge, our work is the first attempt in T2I to incorporate language structures into the image synthesizing process. 

\paragraph{Diffusion Guidance} To convert an unconditional diffusion model into a class-conditional one, \citet{dhariwal2021diffusion} input the noisy image from each step into a classifier and calculate the classification loss. The loss can be back-propagated to the image space to provide a gradient that marginalizes the score estimation from the log of conditional probability. Similarly, in the T2I subdomain, \citet{liu2021more} and \citet{glide} apply a noisy CLIP model to measure the cosine similarity between text prompts and noisy images.

\section{Implementation Details}
Throughout the experiments, we implement our method upon Stable Diffusion v1.4. For all comparisons between our method and Stable Diffusion, we fix the seed to generate the same initial Gaussian map and use 50 diffusion steps with PLMS sampling \citep{liu2021pseudo}. We fix the guidance scale to 7.5 and equally weight the key-value matrices in cross-attention layers if not otherwise specified. We do not add hand-crafted prompts such as ``a photo of'' to the text input. We use the Stanza Library \citep{qi2020stanza} for constituency parsing and obtain noun phrases if not otherwise specified.

\section{\red{Visualization of Attention Maps}}\label{sec:app_vis}
\begin{figure*}[t]
    \centering
    \includegraphics[width=\textwidth]{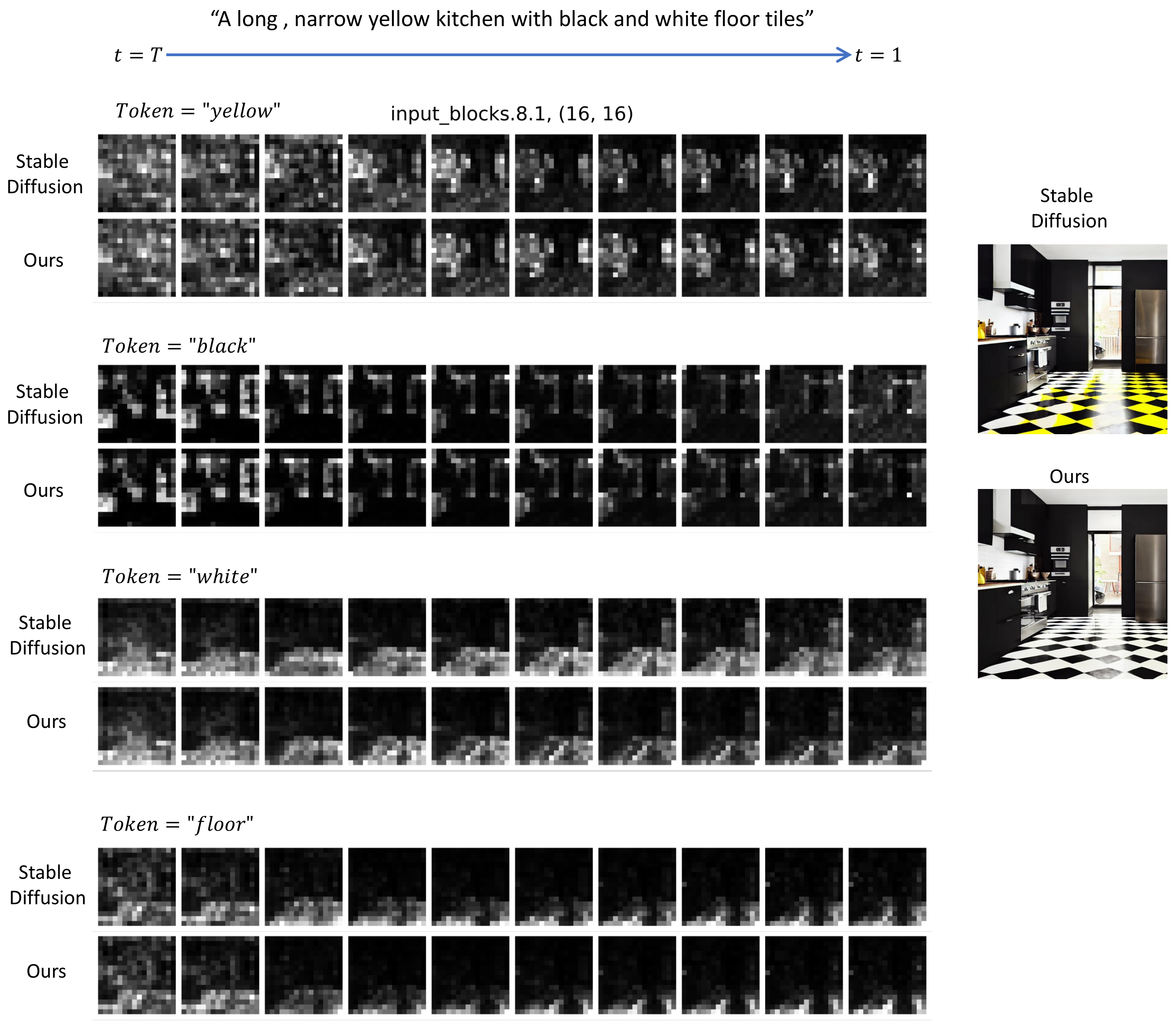}
    \caption{\red{Visualization of cross attention maps of Stable Diffusion and our method. We compare maps of multiple tokens throughout the whole diffusion process with equal intervals.}} 
    \label{fig:map1}
\end{figure*}

\begin{figure*}[t]
    \centering
    \includegraphics[width=\textwidth]{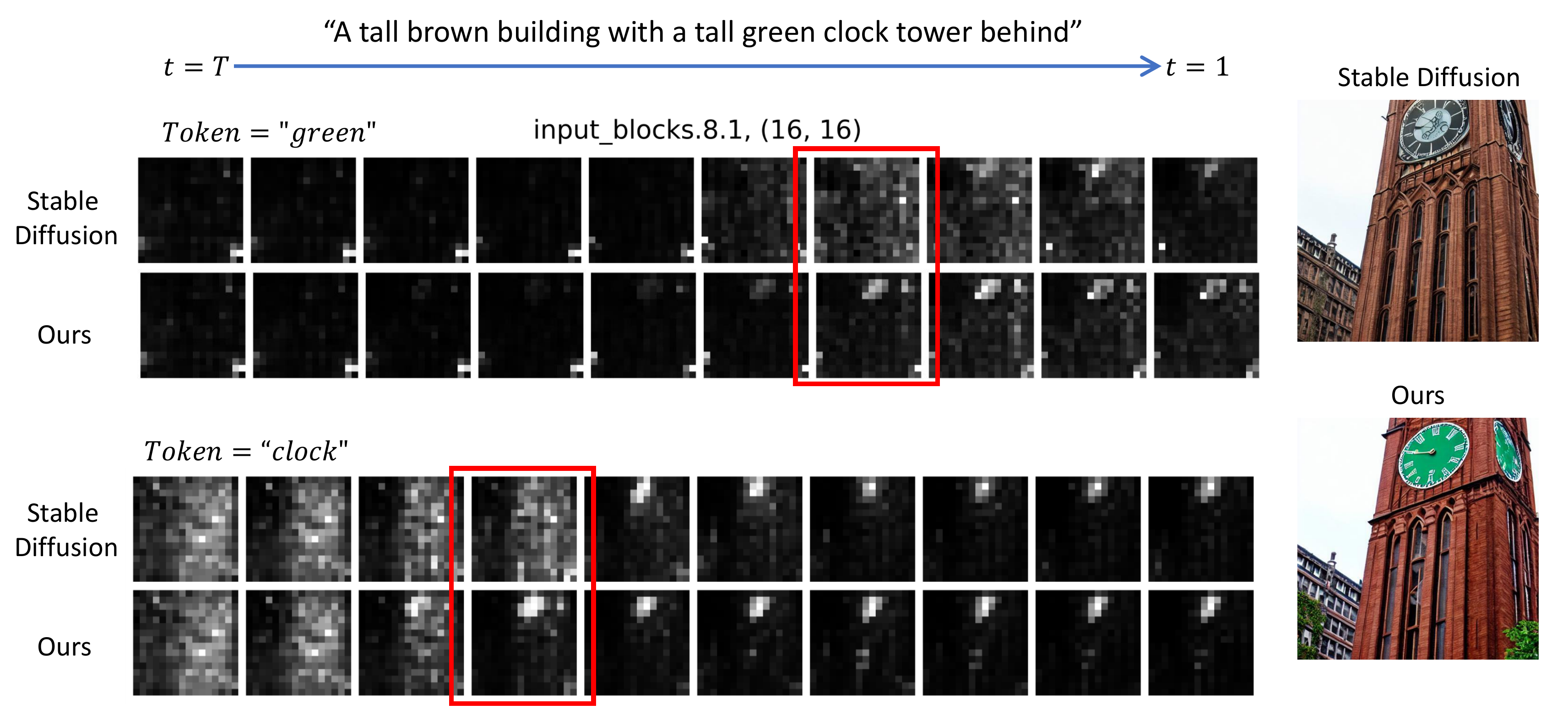}
    \caption{
    \red{Visualization of cross attention maps corresponding to token ``green'' and ``clock'' across the full diffusion timestamps from step 50 to step 1 in equal intervals. Red boxes highlight steps where our method accelerates the formation of correct attention on the clock region. The evolution of the token ``green'' is also more interpretable in our method. Although the image composition is imperfect, the visualization still supports our assumptions and claims in Sec. \ref{subsec:m2}. 
    }}
    \label{fig:map2}
\end{figure*}

\red{In this section, we demonstrate the visualization of cross-attention maps to support our assumptions and claims in Sec. \ref{sec:method}. As is shown in Fig. \ref{fig:map1}, the attention maps of Stable Diffusion and our method have similar spatial distribution and highlights throughout the diffusion process. This phenomenon supports our assumption in Sec. \ref{subsec:m2} that the attention map $M_t$ is unchanged even with multiple values in each cross-attention layer. We can observe a similar phenomenon in Fig. \ref{fig:map2} except that our method accelerates the formation of interpretable attentions for both ``green'' and ``clock'' tokens.}

\begin{figure*}[t]
    \centering
    \includegraphics[width=\textwidth]{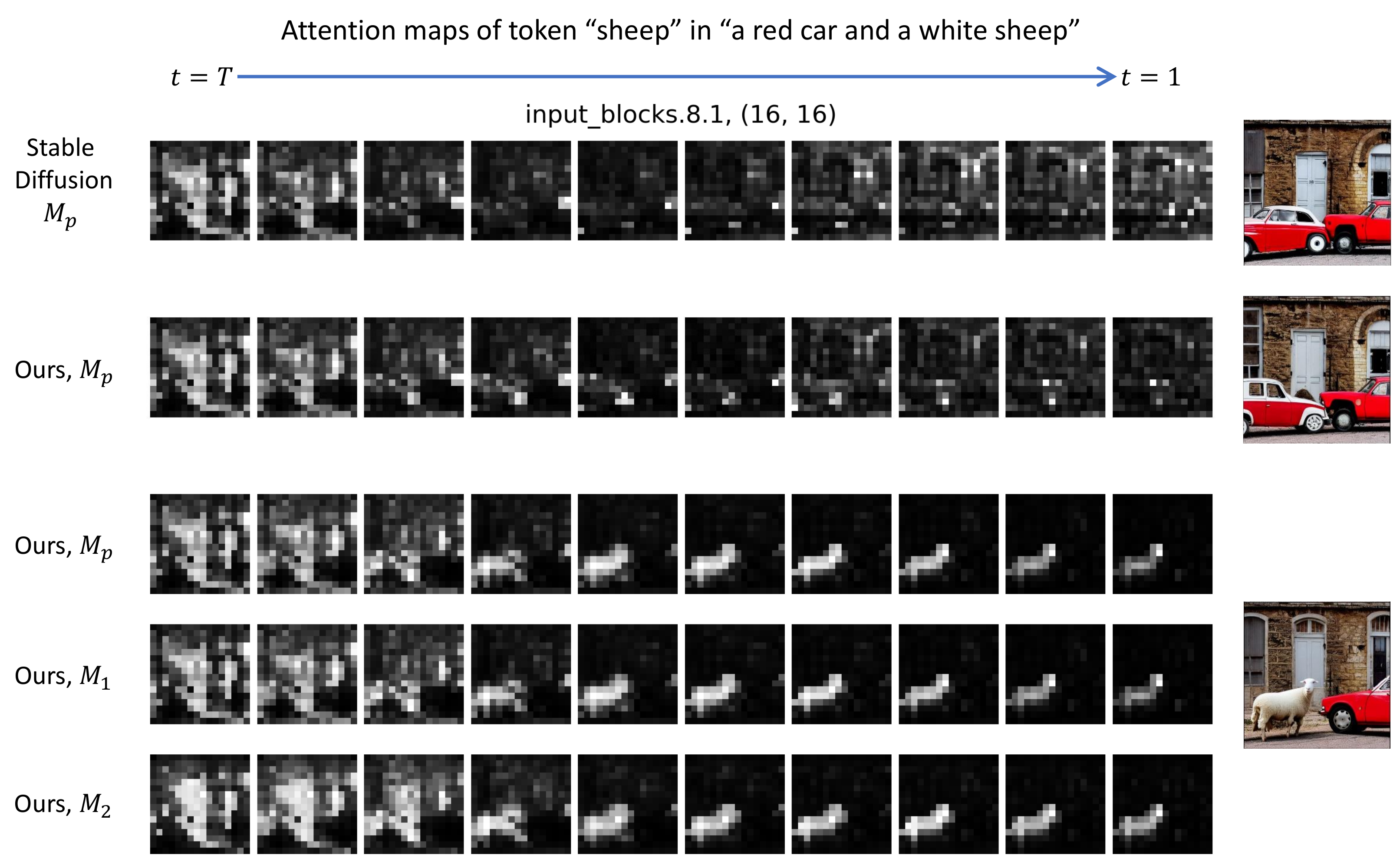}
    \caption{
    \red{Visualization of attention maps for token ``sheep'' of different methods. Our method with multiple Keys successfully rectify image layouts.}}
    \label{fig:map3}
\end{figure*}

\red{Fig. \ref{fig:map1}, \ref{fig:map2} also justify our claim that values represent rich textual semantics mapped to the image space as contents. For instance, our method parses the prompt in Fig. \ref{fig:map1} into ``A long narrow yellow kitchen'' and ``black and white floor tiles'', encodes and aligns them separately to form $\mathbb{V}$. Empirically, these operations enhance the semantics of ``yellow'' and ``black and white'' separately and mitigate ``yellow'' being blended into ``black and white''. This explains the disappearance of color leakage in our image compared to Stable Diffusion. Though one may attribute the leakage to incorrect attention distribution of the ``yellow'' token, we argue that this is not the critical reason. Despite the attention maps of ``yellow'' from our method slightly highlighting the ``floor tile'' regions, we cannot observe any yellow in our generated image. This proves that inaccurate attention distributions contribute little to the final image content.} \red{In addition, we also show in Fig. \ref{fig:map3} that using multiple Keys is able to rectify the image layouts to mitigate missing object issues. The sheep-like attention maps in the third row verify the proposed variants of our method for concept conjunctions.}

\section{Ablation Study}\label{sec:app_ablation}

\subsection{A case study of attribute binding}
\begin{figure*}[t]
    \centering
    \includegraphics[width=\textwidth]{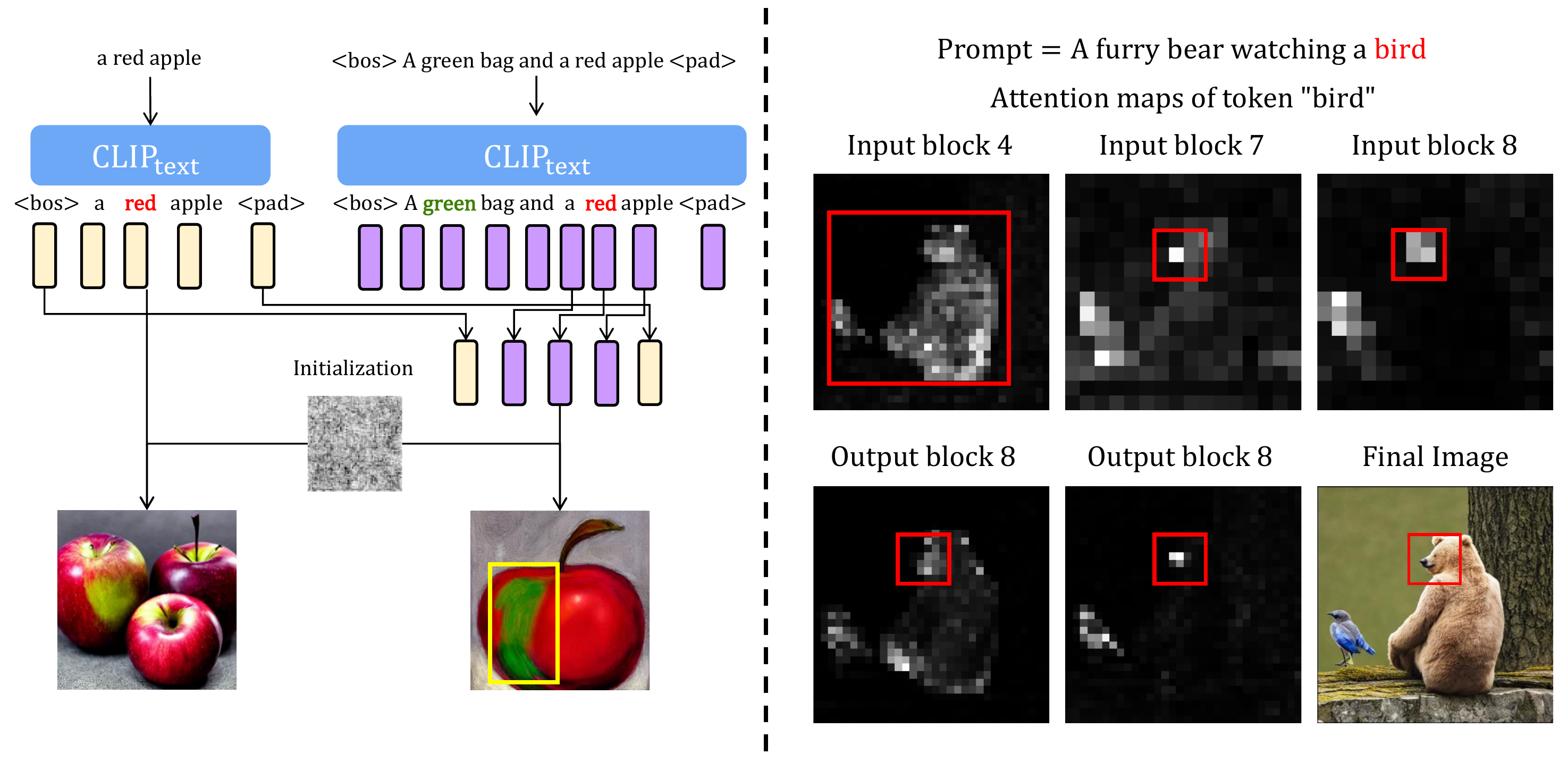}
    \caption{Examples showing the potential root causes of incorrect attribute binding. \textbf{Left}: The large green regions in the second image prove that the hidden state's output of token ``red'' is contextualized with token ``green'' before it. \textbf{Right}: Visualization of attention maps showing that the semantics from the token ``bird'' is mistakenly attended to the mouth region of the bear. The final image shows the unnatural beak-like shape of the bear.}
    \label{fig:case_study}
\end{figure*}

Here, we present a case study to show evidence of two root causes of incorrect attribute binding. The first one is the contextualized token embeddings due to causal attention masks. As is shown on the left side of Fig. \ref{fig:case_study}, we first encode two different prompts with a shared component, e.g. ``a red apple'' as the naive one and ``a green bag and a red apple''. Using the encoding sequence of the naive prompt, we are able to get an image of red apple only. It is reasonable to assume that the yellow \\ green regions are natural results of learning from authentic apple images. Then, we replace the tokens of the naive prompt with embeddings of the same token from the more complicated prompt. We use the same gaussian noise as initialization and generate an unnatural image with a solid green region (in the yellow bounding box). This result proves that the token ``red'' is contaminated with the semantics of ``green'' before it and explains some images with color leakage problems (e.g., Fig. \ref{fig:teaser}).

The second reason attributes to inaccurate attention maps. In Fig. \ref{fig:case_study} (right), we visualize five cross-attention maps (averaged across attention heads) from both downsample and upsampling blocks. The attention maps show the salient regions corresponding to the token ``bird''. These maps demonstrate highlighted regions in the bottom left corner where the bird is located in the final image. Despite the interpretable structures, the maps also show saliency around the mouth region of the bear across all five layers. Thus, the inaccurate attention maps lead to a beak-like mouth of the bear in the image.

\subsection{Comparison of Parsers} \label{subsec:parsers}
\begin{figure*}[t]
    \centering
    \includegraphics[width=\textwidth]{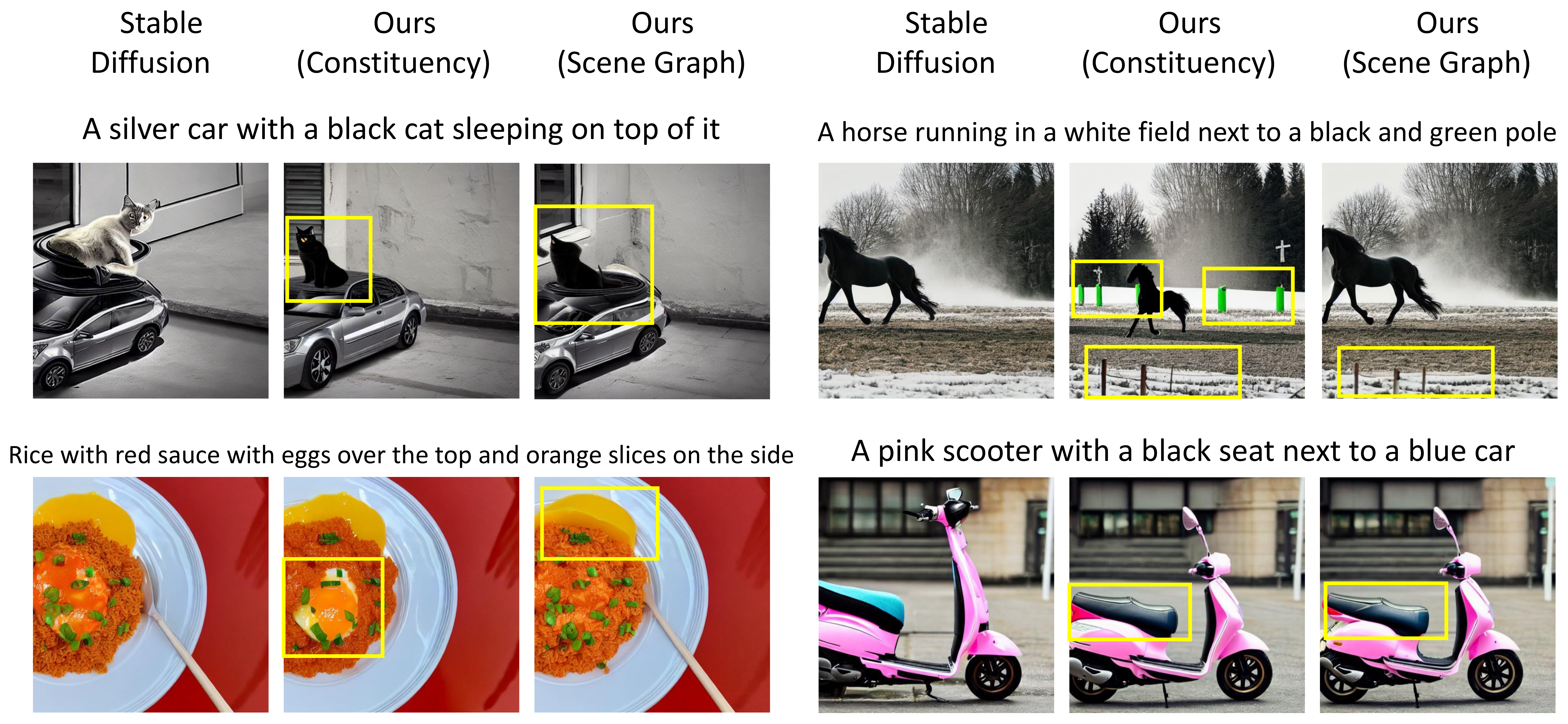}
    \caption{\red{Synthesized images corresponding to prompts in Table \ref{tab:parsers}. Yellow boxes annotate compositions that are improved using different parsers.}}
    \label{fig:sg}
\end{figure*}

\red{In this subsection, we compare the difference between using a constituency parser and a scene graph parser to obtain text spans and generate images. Table \ref{tab:parsers} compares the extracted text spans using constituency parser and scene graph parser. Example 0 shows that both parsers end up with the same results for CC-500 prompts. For Example 1-4, the scene graph parser generates more spans than the constituency parser. We notice that concepts in the middle of the sentence appear more often in these spans than other noun tokens, like ``egg'' or ``red sauce'' in Example 3. This imbalance potentially explains why the ``egg'' looks more highlighted in Fig. \ref{fig:sg} (bottom left). On the other hand, ``orange slices'' appear more often in constituency parsing results, leading to better ``orange'' textures in the generated image. Similar observations can be made in Example 2, where ``green pole'' is emphasized more often by the constituency parser. 
}

\begin{table*}[t]
\centering
\resizebox{\textwidth}{!}{%    
\begin{tabular}{l|c|c}
\toprule
\midrule
& \multirow{1}{*}{\textbf{Constituency Parser}} & \multirow{1}{*}{\textbf{Scene Graph Parser}} \\
    % Example 0
    \midrule
    \multirow{2}{*}{Example 0} & \multicolumn{2}{c}{CC-500 Prompt: \textit{A white sheep and a red car}}   \\
    \cmidrule{2-3}
    & \makecell{``A white sheep'', ``a red car''} 
    & \makecell{``A white sheep'', ``a red car''} \\
    % Example 1
    \midrule
    \multirow{2}{*}[-1em]{Example 1} & \multicolumn{2}{c}{Prompt: \textit{A silver car with a black cat sleeping on top of it}}   \\
    \cmidrule{2-3}
     & \makecell{``A silver car'', ``a black cat'', \\ ``A silver car with a black cat''}
     & \makecell{``A silver car'', ``a black cat'', ``top of it'',\\ ``a black cat sleeping on top of it''} \\
    %  Example 2
    \midrule
    \multirow{2}{*}[-2em]{Example 2} & \multicolumn{2}{c}{Prompt: \textit{A horse running in a white field next to a black and green pole}}   \\
    \cmidrule{2-3}
     & \makecell{``A horse'', ``a white field'', \\ ``a black and green pole'',\\ ``a white field next to a black and green pole''}
     & \makecell{``A horse'', ``a white field'',\\ ``a black and green pole'',\\ ``A horse running in a white field''} \\
    %  Example 3
    \midrule
    \multirow{2}{*}[-3em]{Example 3} & \multicolumn{2}{c}{Prompt: \textit{Rice with red sauce with eggs over the top and orange slices on the side}}   \\
    \cmidrule{2-3}
     & \makecell{``red sauce'', ``the side'', \\ ``the top and orange slices'', \\ ``the top and orange slices on the side''}
     & \makecell{``red sauce'', ``the side'',\\ ``the top and orange slices'',\\  ``Rice with red sauce'', ``red sauce with eggs'',\\ ``the top and orange slices on the side'', \\ ``red sauce with eggs over the top and orange slices''} \\
    %  Example 4
     \midrule
     \multirow{2}{*}[-1.5em]{Example 4} & \multicolumn{2}{c}{Prompt: \textit{A pink scooter with a black seat next to a blue car}}   \\
    \cmidrule{2-3}
    & \makecell{``A pink scooter'', ``a black seat'', ``a blue car''} 
    & \makecell{``A pink scooter'', ``a black seat'', ``a blue car'', \\``a pink scooter with a black seat'', \\``a black seat next to a blue car''} \\
    \midrule
    \bottomrule
\end{tabular}
}
\caption{\red{Comparison between the constituency parser and scene graph parser. For CC-500 prompts, both parsers end up with the same results. As for general prompts, scene graph parser tends to generate more text spans with middle concepts appearing multiple times across different spans.}}
\label{tab:parsers}
\end{table*}

\section{Limitations \& Future Work} \label{sec:limits}
There are several limitations of our work. First of all, our method depends on an external parsing function that may not be perfect. We adopt the commonly used Stanza Library \cite{qi2020stanza} for constituency parsing. The parsing function can be replaced with a more advanced learning-based method for improvement. Secondly, our method mainly focuses on compositional T2I neglecting any style descriptions. The parsing mechanism may categorize a style description, e.g. ``in Van Gogh style'' as a separate noun phrase that cannot be grounded in the image space. In addition, we discover that StructureDiffusion tends to generate similar images as Stable Diffusion. Thus we filtered out 20\% of most similar image pairs in Table \ref{tab:human}, considering the efficiency of human evaluation. Therefore, the improvement could be compromised when evaluated on the full set of generated images. Future work may focus on devising explicit methods to associate attributes to objects using spatial information as input. For example, how to make a text-to-image synthesis model interpret coordinate information with limited fine-tuning or prompt tuning steps would be an appealing direction.

\section{Additional Results}

\begin{figure*}[t]
    \centering
    \includegraphics[width=\textwidth]{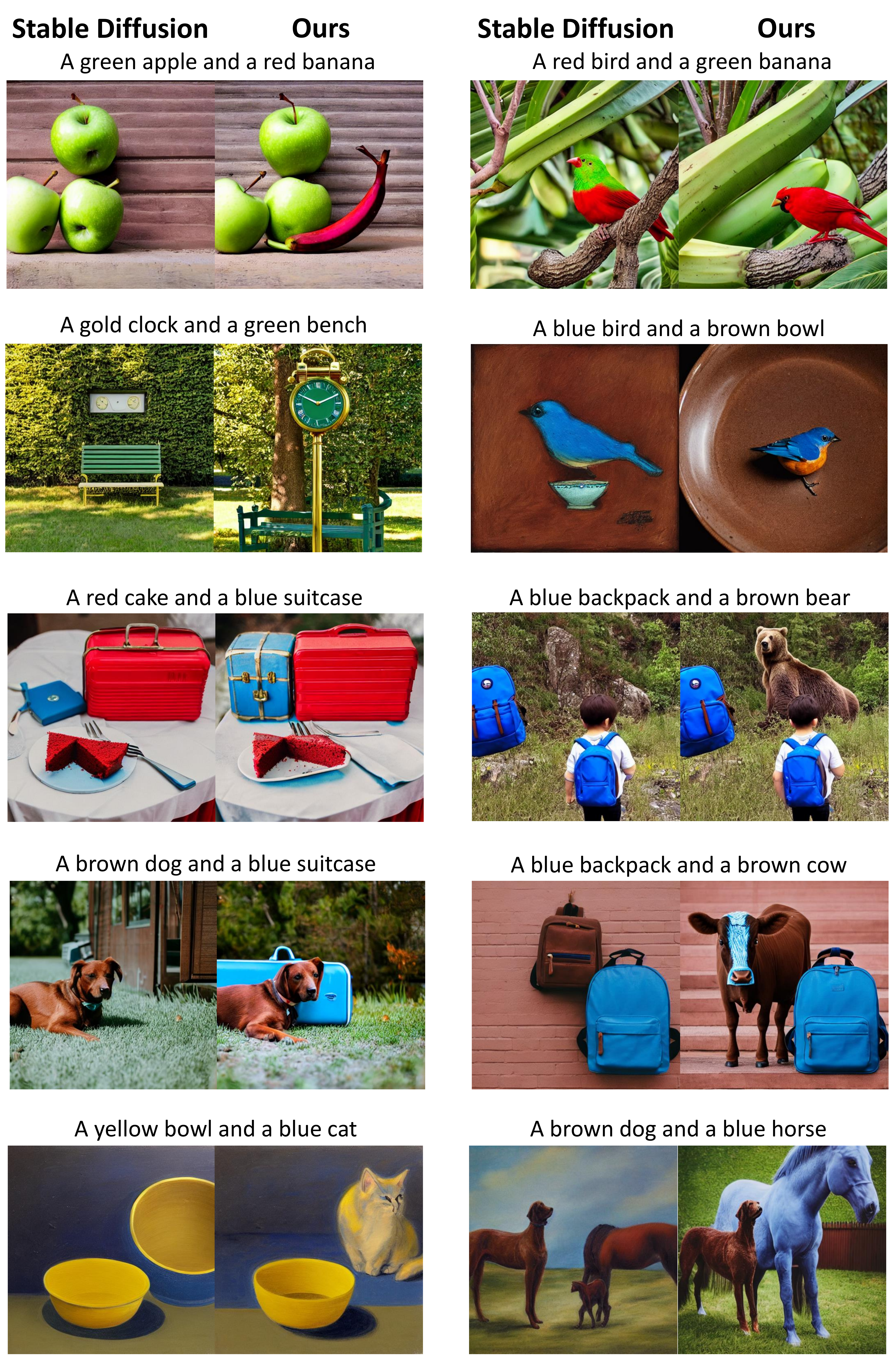}
    \caption{Qualitative results on CC-500
    }
    \label{fig:cb_appendix}
\end{figure*}

\begin{figure*}[t]
    \centering
    \includegraphics[width=\textwidth]{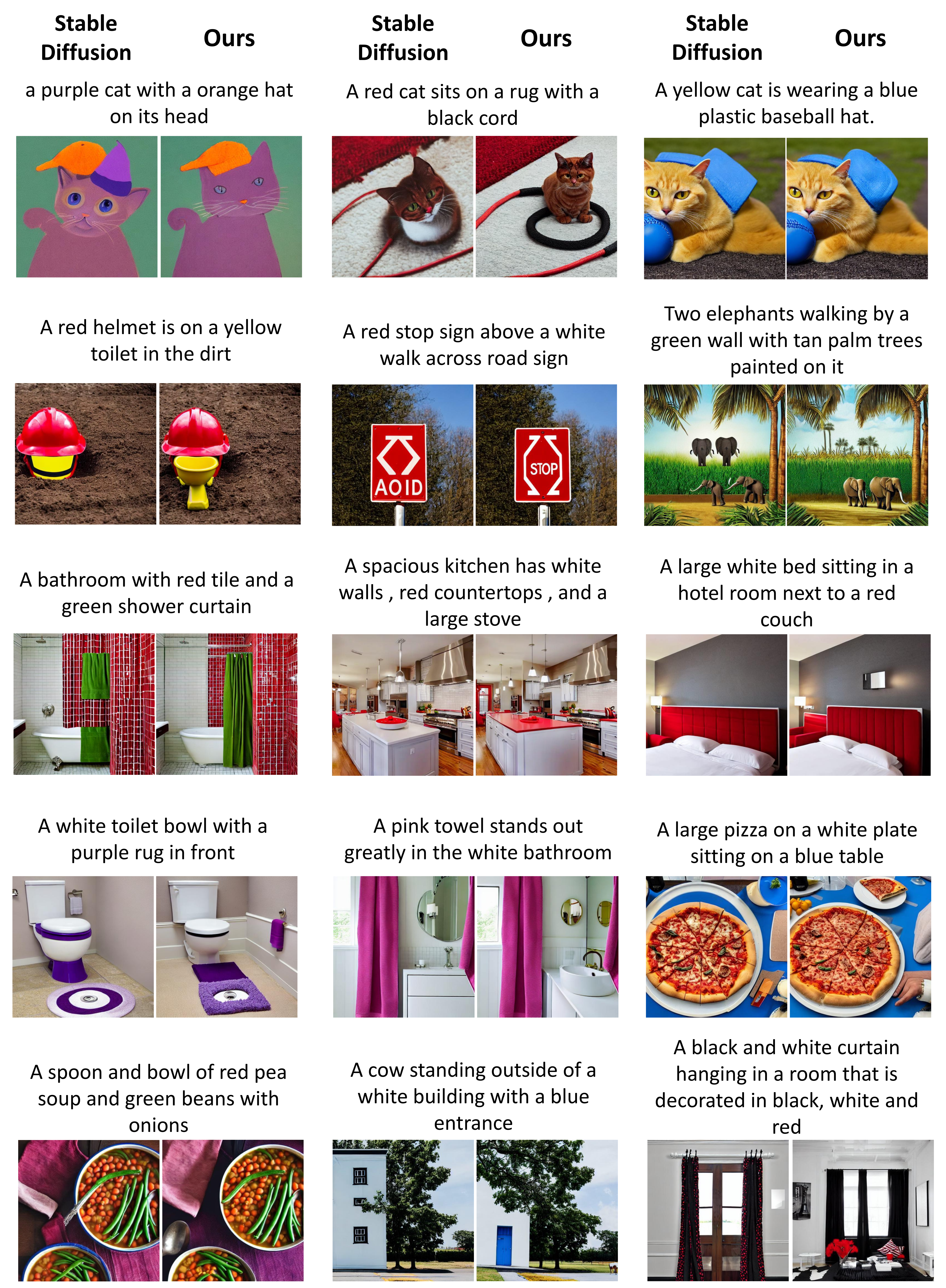}
    \caption{\red{Qualitative results on ABC-6K}}
    \label{fig:abc_appendix}
\end{figure*}

\begin{figure*}[t]
    \centering
    \includegraphics[width=\textwidth]{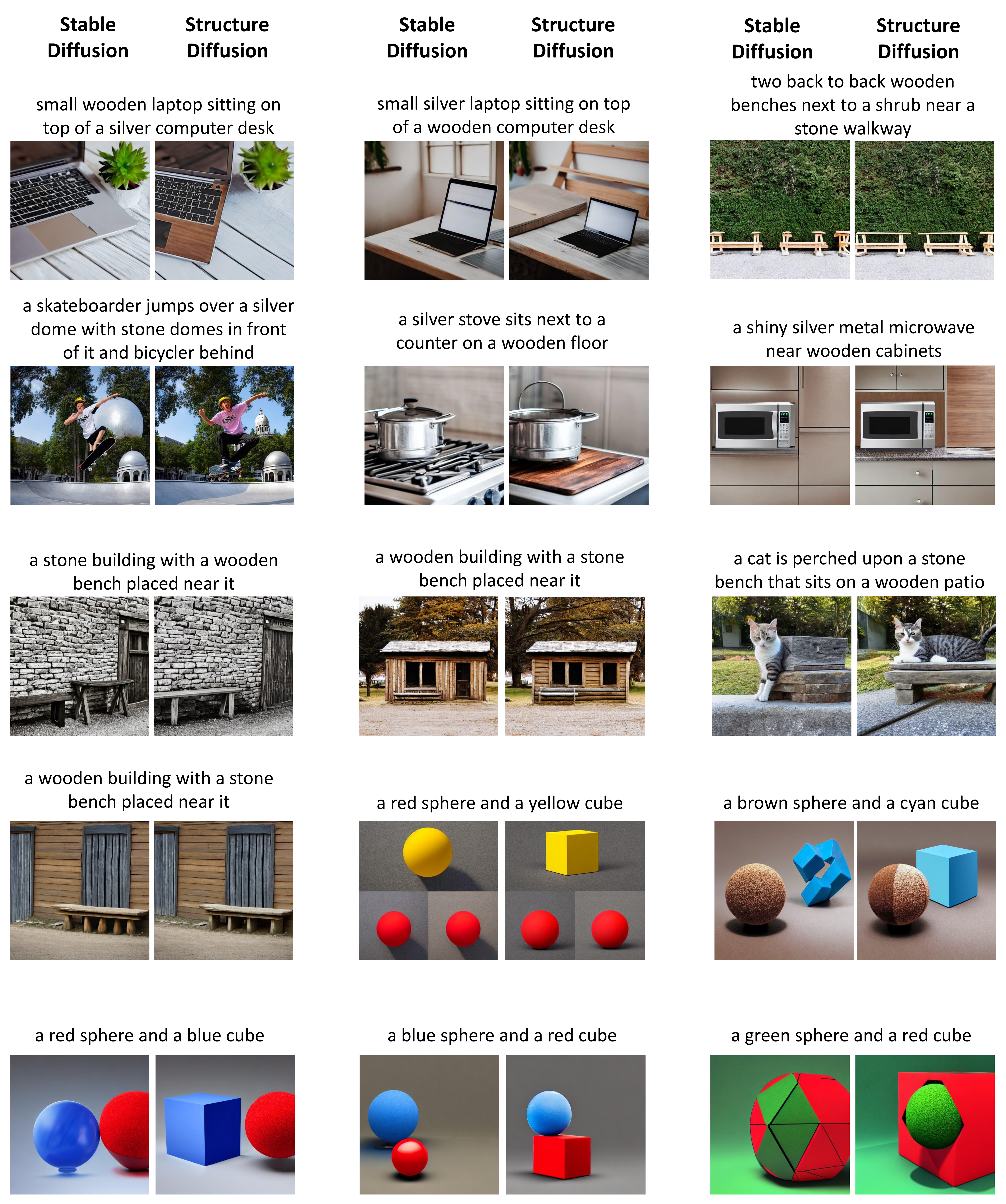}
    \caption{\red{Qualitative results characterizing attributes beyond colors, including shape, size and materials.}
    }
    \label{fig:attri_appendix}
\end{figure*}

\begin{figure*}[t]
    \centering
    \includegraphics[width=\textwidth]{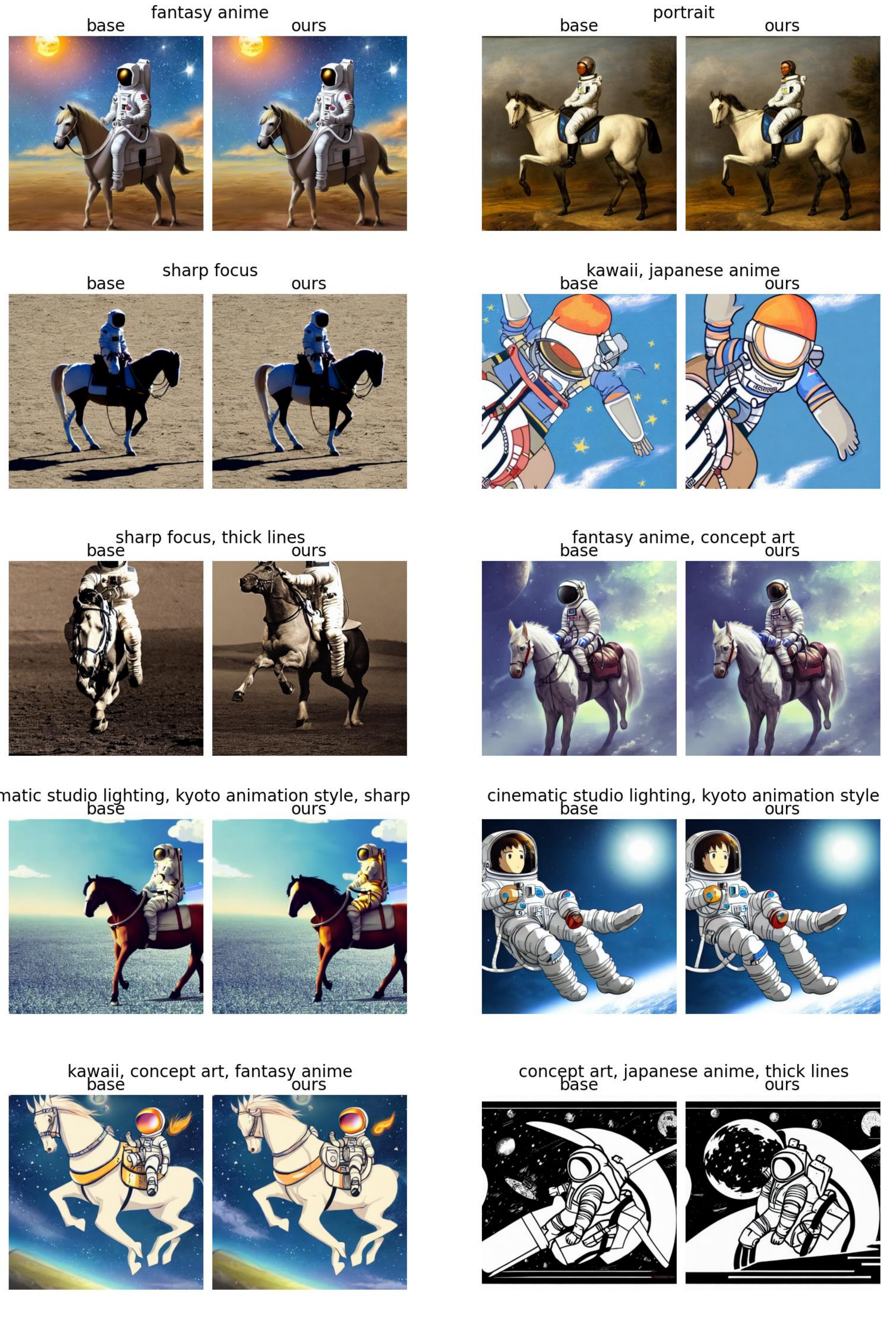}
    \caption{\red{A prompt ``an astronaut riding a horse'' appended with different (combinations of) style descriptions. Our method has no negative effects on the image style. ``base'' refers to Stable Diffusion.}}
    \label{fig:style}
\end{figure*}
% You may include other additional sections here.

\end{document}